\documentclass[10pt,twocolumn,letterpaper]{article}

\usepackage[pagenumbers]{cvpr} 

\usepackage{graphicx}
\usepackage{amsmath}
\usepackage{amssymb}
\usepackage{booktabs}

\usepackage[utf8]{inputenc} 
\usepackage[T1]{fontenc}    
\usepackage{url}            
\usepackage{booktabs}       
\usepackage{amsfonts}       
\usepackage{nicefrac}       
\usepackage{microtype}      
\usepackage[table]{xcolor}         

\usepackage{amsmath}
\usepackage[numbers]{natbib}
\usepackage{xspace}
\usepackage{graphicx}
\usepackage[export]{adjustbox}
\usepackage{algorithm}
\usepackage[noend]{algpseudocode}
\usepackage{verbatim}
\usepackage{subcaption}
\usepackage{multirow}
\usepackage{hhline}
\usepackage{pifont}  
\usepackage{wrapfig}

\usepackage{tabularx}

\usepackage[pagebackref,breaklinks,colorlinks]{hyperref}
\usepackage{cleveref}
\crefname{section}{Sec.}{Secs.}
\Crefname{section}{Section}{Sections}
\Crefname{table}{Table}{Tables}
\crefname{table}{Tab.}{Tabs.}

\crefformat{section}{\S#2#1#3}
\crefformat{subsection}{\S#2#1#3}
\crefformat{subsubsection}{\S#2#1#3}
\crefrangeformat{section}{\S\S#3#1#4 to~#5#2#6}
\crefmultiformat{section}{\S\S#2#1#3}{ and~#2#1#3}{, #2#1#3}{ and~#2#1#3}
\usepackage{refstyle}
\Crefformat{figure}{#2Fig.~#1#3}
\Crefmultiformat{figure}{Figs.~#2#1#3}{ and~#2#1#3}{, #2#1#3}{ and~#2#1#3}
\Crefformat{table}{#2Tab.~#1#3}
\Crefmultiformat{table}{Tabs.~#2#1#3}{ and~#2#1#3}{, #2#1#3}{ and~#2#1#3}
\Crefformat{appendix}{Appx.~\S#2#1#3}
\Crefformat{section}{Sec.~\S#2#1#3}
\crefformat{algorithm}{Alg.~#2#1#3}
\usepackage{xcolor}

\newcommand{\xmark}{\textcolor{red}{\ding{55}}}
\newcommand{\cmark}{\textcolor{blue}{\ding{51}}}

\def\cvprPaperID{10353} 
\def\confName{CVPR}
\def\confYear{2023}

\begin{document}







\title{DALL-E for Detection: Language-driven Compositional Image Synthesis for Object Detection}


\author{Yunhao Ge$^{1*}$, Jiashu Xu$^{1*}$, Brian Nlong Zhao$^{1}$, Laurent Itti$^{1}$, Vibhav Vineet$^{2}$\\
$^{1}$University of Southern California \ \ \  $^{2}$Microsoft Research\ \ \ \
{\tt\small $^{*}$co-first author}}
\maketitle

\begin{abstract}
We propose a new paradigm to automatically generate training data with accurate labels at scale using the text-to-image synthesis frameworks (e.g., DALLE, Stable Diffusion, etc.). 
The proposed approach decouples training data generation into foreground object mask generation and background (context) image generation. 
%
For foreground object mask generation, we use a simple textual template with object class name as input to DALLE to generate a diverse set of foreground images. A foreground-background segmentation algorithm is then used to generate foreground object masks. 
Next, in order to generate context images, first a language description of the context is generated by applying an image captioning method on a small set of images representing the context. These language descriptions are then used to generate diverse sets of context images using the DALL-E framework. These are then composited with object masks generated in the first step to provide an augmented training set for a classifier.
We demonstrate the advantages of our approach 
on four object detection datasets including on Pascal VOC and COCO object detection tasks.
Furthermore, we also highlight the compositional nature of our data generation approach on out-of-distribution and zero-shot data generation scenarios.
%
\end{abstract}

\section{Introduction}
\label{sec:intro}


Training modern deep learning models require large labeled datasets \cite{ren2017faster,he2016deep,Shelhamer_2017}. Obtaining such datasets is both expensive and time-consuming due to large human effort requirements.
%
This raises a question: can we efficiently generate a large scale labelled data that also achieves high accuracy on a new downstream task? We first hypothesize that any such approach should satisfy these qualities (Table.~\ref{table:advantage}): no (or minimal) human involvement, automatic generalization of the images for any new classes and environments, scalable, generation of high quality and diverse set of images, explainable, compositional, and privacy-preserving.

To this end, synthetic techniques could be used as possible sources for generating labelled data for training computer vision models.
%
%
One popular approach is to use computer graphics to generate data~\cite{richter2016playing,song2016ssc,FallingThings}. However, these approaches may require gathering 3D models of both objects and scenes, which can require a large amount of skilled labor, such as 3D modeling expertise, which prohibits the scalability of such graphics generated synthetic labelled data.
Another approach is to use object-cut and paste~\cite{dwibedi2017cut}, which is a 2D synthetic generation approach, but these approach can have limitations as they still require a source for foreground objects and accurate foreground masks for those objects.
A third approach could be to use machine learning based neural renderer techniques like NeRF based approaches \cite{mildenhall2020nerf,NeuralSim}. These approaches generally require retraining models for any new object class and so can not easily scale to large number of object classes.

\begin{figure}[t]
\begin{center}
\includegraphics[width=\linewidth]{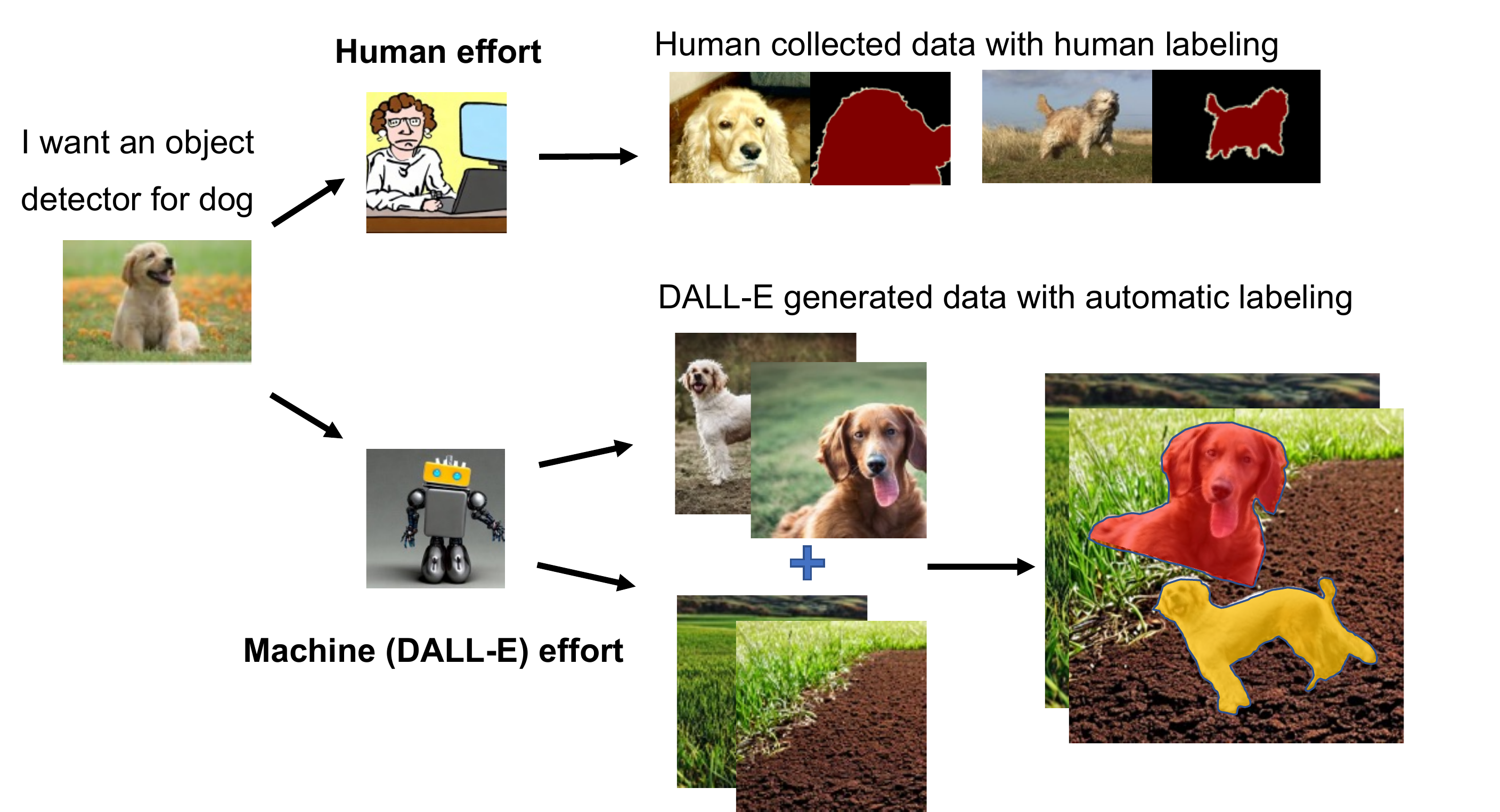}
\end{center}
   \caption{ Comparison of DALL-E for detection pipeline and traditional human-centric pipeline.
   }
\label{fig:teaser}
\end{figure}
Recently, there has been a revolution in large scale text-to-image synthesis models like DALL-E \cite{ramesh2021zero}, RU-DALLE \cite{sberbank31:rudalle}, CogView \cite{ding2021cogview}, and Stable Diffusion \cite{rombach2022high}. 
They have been shown to achieve photorealism even for complex scenes, able to  understand semantics and compositional nature of the real world. In addition, these models can understand language descriptions of a scene 
and so can act a natural bridge between human and synthesis approaches. 
Given these qualities, can these text-to-image synthesis approaches be used to generate large scale training data with accurate labels for computer vision problems? For simplicity, from now on, we will use DALL-E to represent text-to-image synthesize models (including Stable Diffusion and RU-DALLE).

\begin{table*}
\begin{center}
\begin{tabular}{c|c c c c c c c}
\hline
    Method  & Quality & No Human & Adapt & Scalable  & Explainable & Privacy & Comp. \\
\hline
  Human capture & \cmark & \xmark & \xmark & \xmark & \cmark & \xmark & \cmark \\ 
  Web image    & \cmark & \cmark & \xmark & \cmark & \xmark & \xmark  & \xmark \\ 
  Public dataset  & \cmark & \xmark & \xmark & \cmark & \xmark & \xmark  & \xmark \\ 
  Generative models   & \cmark & \cmark & \xmark & \cmark & \xmark & \cmark & \xmark \\  
  Ours & \cmark & \cmark & \cmark & \cmark & \cmark & \cmark & \cmark \\
\hline
\end{tabular}
\end{center}
\vspace{-5mm}
\caption{Desired quality of context generation method: Generation of a high quality and diverse set of images (scalable), no(less) human involvement in context image generation, automatic generalization of the images for any new environment,  scalable, explainable, privacy-preserving, and compositional.
} 
\label{table:advantage}
\end{table*}

In this work, we propose a new text-to-image synthesis paradigm involving two components to generate large scale training data with accurate labels for tasks like object detection and instance segmentation. 
First part involves accurately generating foreground object masks for the object classes of interest. This is due to the reason that DALL-E can not generate annotations, e.g., they are not able to generate bounding box for the objects. 
In order to generate diverse foreground object masks, we first generate images containing mostly one object corresponding to  the interest class. We use a simple template with class name as input to the DALL-E pipeline. For example, in order to generate images with {\em cat}, we use input as {\em an image of cat on pure background.} Next, a simple background-foreground segmentation approach is used to get foreground object masks of the interest classes. 

The second step involves generating diverse background images that provide good context information for training recognition models. 
%
Context plays an important role in learning a good object recognition model. Divalla et al. \cite{divvala2009empirical} provide  empirical evidence to support this claim. Dvornik \cite{dvornik2018modeling} showed that finding congruent context helped improve accuracy on object detection tasks. For example, placing airplanes and boats in their natural context helped to improve accuracy, e.g., airplanes are generally found in the sky and boats are on the water. 

We again leverage DALL-E to generate diverse set of high quality context images. At the core of our approach lies utilizing an interplay between language description of context and language-driven image generation. %
Given a small number of images that represent the context environment, we use image captioning  to generate a high-level language description of the context automatically. The language description of the context is used within a text-to-image generation pipeline (e.g., DALL-E) to generate a diverse set of images. 
These diverse sets of generated images are used as context images. 


Finally, to generate labelled data, we follow a simple strategy where we paste the foreground object masks (from step one) onto the random context images (from step two), as in object-cut and paste approaches~\cite{dwibedi2017cut}. The proposed pipeline satisfies all the desired properties of labelled image generation (Fig.~\ref{fig:capgen}). The data can be generated efficiently without human involvement effortlessly. 
Language descriptions for both the foreground and the background image generation help to provide an explainable and compositional data generation. Adding or removing objects or settings can be easily done in the language domain. For example, a description as {\em an environment with a table} can be easily modified to kitchen environment by utilizing the compositional properties of language as {\em a kitchen environment with a table}. 
Table. \ref{table:advantage} shows the benefit of our approach in generating context images over other approaches. 

We have conducted extensive experiments on four publicly available benchmark object detection datasets and compared them against different ways to generate context images. We demonstrate that our approach can achieve much better accuracy compared to the prior approaches. We also demonstrate the benefit of our approaches in out-of-distribution context and zero-shot data generation  scenarios that utilize the compositional nature of our method.
Our main contributions are:
(1) We propose a language-driven compositional image generation approach to automatically generate both foreground objects and background context images and form large-scale datasets.
(2) We demonstrate benefit of our proposed pipeline over several prior approaches. 
(3) We highlight compositional nature by generating context images from out-of-context images and zero-shot data generation scenarios.  To the best of our knowledge, this is the first work to use vision and language models for generating object detection and segmentation datasets. 



\section{Related works}
\label{sec:related_works}

\paragraph{Text-to-Image Synthesis Frameworks.} 
\vspace{-2mm}
Text-to-image synthesis approaches have become new revolution in generating high-quality images, capturing semantics and compositionality of the real world scenes. Some of these approaches include DALL-E \cite{ramesh2021zero}, RU-DALLE \cite{sberbank31:rudalle}, Stable Diffusion \cite{rombach2022high}, CogView \cite{ding2021cogview}. These approaches leverage benefits of using large scale transformer based models trained using large scale vision and text data. Though they have shown to generate high quality images of complex real world scenes, they are not able to generate ground truth labels for objects. For example, they are not able to provide bounding box and per-pixel annotation for objects of interest. In our work, we have proposed an automatic approach to generate high quality images with ground truth bounding box and per-pixel labels.
\vspace{-5mm}
\paragraph{Synthetic Data Generation.} 
A series of works on using synthetic data for training computer vision problems have been proposed. Some of them include using graphics pipeline or computer games to generate high quality labelled data \cite{richter2016playing,RichterHK_iccv17,RosSMVL_cvpr16,hodan2019photorealistic,TremblayTB_corr18,HandaPBSC_corr15a}. Generally using graphics pipeline requires having 3D models of both objects and environment, that may limit their scalability. Some of them use generative models (e.g., GAN) \cite{bowles2018gan,ge2020pose} or zero-shot synthesis \cite{ge2020zero} to augment datasets and remove bias. However, they need a relatively large initial dataset to train the model, and not easy to generalize to new domains.

The idea of pasting foreground objects on background images has emerged as a easy and scalable approach for large scale data generation. 
%
%
The idea has been used to solve other problems like object instance segmentation tasks \cite{dwibedi2017cut}, object detection and pose estimation problems \cite{SuQLG15,hinterstoisser2017pre,rad2017bb8,TekinSF_corr17,tripathi2019learning}, optical flow \cite{Dosovitskiy_2015_ICCV}, domain adaptation problem \cite{yun2021cut}, semi-supervised learning task \cite{ghiasi2021simple}. 
%
%
These approaches generally require accurate foreground object masks. This limits their scalability. While we utilize a cut-and-paste approach, in contrast to previous works in this space, our work can generate foreground object masks for any new object class.

\vspace{-5mm}
\paragraph{Language for object recognition.} Language has been used to solve computer vision tasks. Vision-language based models have been developed for image captioning task \cite{vinyals2015show,rennie2017self,anderson2018bottom,gurari2020captioning}, visual question answering tasks \cite{antol2015vqa,vedantam2019probabilistic,agrawal2018don,das2018embodied} and others  \cite{lu202012,yang2021causal,li2020unicoder}. In recent years, vision and language based multi-modal models have been developed for self-supervised training. 
Recent work CLIP \cite{radford2021learning} showed how training a model on large image-text pair dataset can generalize to several benchmark image classification datasets where current image based models performed very poorly. Language information been used along with vision information to solve other computer vision tasks like object detection \cite{gu2021zero,kamath2021mdetr,li2022grounded} and semantic image segmentation task \cite{li2022language}. These works have also demonstrated benefits of leveraging language information in solving computer vision tasks.

Another line of work involves using large image-text pairs for text to image generation tasks. These models learnt to generate reaslitic environment including out-of-context settings which may not be present during the training phase. Some of the examples include DALL-E \cite{ramesh2021zero}, RU-DALLE \cite{sberbank31:rudalle}, CogView \cite{ding2021cogview} and Stable Diffusion \cite{rombach2022high}. 
A recent concurrent work X-Paste \cite{zhao2022x} has used Stable Diffusion to solve object detection. 
We are motivated by generation quality of these text to image generation methods in this work.

%




\vspace{-2mm}
\section{Method}
\label{sec:method}

\vspace{-1.5mm}

The goal of the paper is to efficiently generate a large set of labeled data for training object detection models using text-to-image synthesis frameworks. 
In particular, the proposed approach decouples training data generation into diverse set of foreground object mask generation and diverse set of background (context) image generation. 
%
%
%
The foreground object masks are then composited onto the background images following the object cut-and-paste strategy \cite{dwibedi2017cut}. 
Furthermore, the proposed approach also allows for compositional and explainable data generation as well. 
%
Our pipeline leverages off-the-shelf
language generative frameworks \cite{ho2020denoising, rombach2022high} under textual guidance \cite{ho2022classifier} to generate both the foreground masks and context images. 

\subsection{Language-driven Context generation} 
\label{sec:syn_gen}
\begin{figure*}
\begin{center}
\vspace{-40pt}
\includegraphics[width=\linewidth]{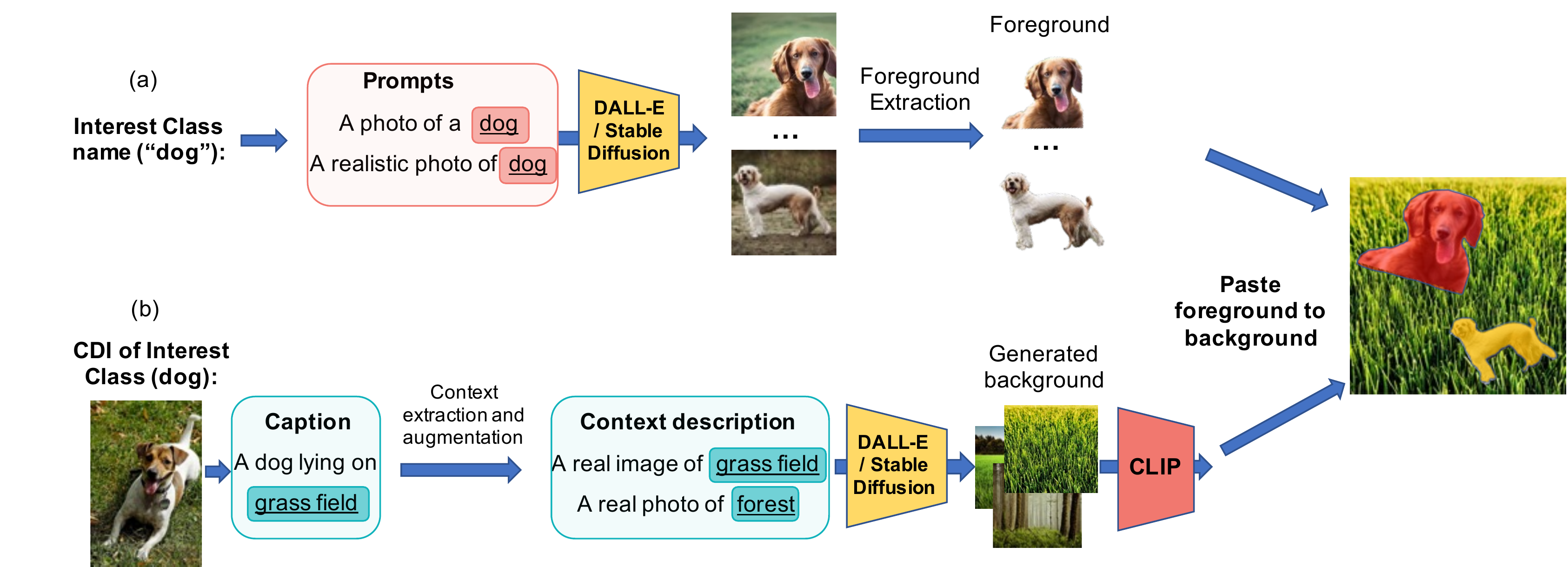}
\end{center}
\vspace{-15pt}
    \caption{ Our pipeline consists of foreground generation and context background generation. (a) Foreground generation (top row): (1) we fill the interest class name (e.g., dog) into fixed prompt templates to produce foreground sentences. (2) We then feed the sentences to DALL-E (or Stable diffusion) to generate high quality foreground images with easy to separate background. 
    (3) We use off-the-shelf image segmentation methods to extract foreground segments from foreground images. 
    (b) Background context generation (bottom row): (4) we use image captioning method (e.g., SCST \cite{rennie2017self}) to generate captions for the user provided CDIs (the user can provide as little as one image). (5) we leverage lexical networks and models to extract the background context words (e.g., grass field) and augment more related context images based on ConceptNet (e.g., forest). (6) We create context description sentences based on the context words with templates. (7) We feed the sentences to DALL-E (or Stable diffusion) generate high quality background images. 
    (8) We use CLIP \cite{radford2021learning} to filter and further ensure that the generated images have no interest class. (9) We combine the foreground segments and background images to obtain synthetic images with corresponding annotations using cut and paste. (10) We use the synthetic dataset to train object detection/segmentation models.}
\label{fig:capgen}
\end{figure*}


Context image generation consists of two steps (a visualization of our pipeline is provided in Fig.\ref{fig:capgen}). The first step involves generating a diverse set of language descriptions of context by applying image captioning methods on few images taken from the context environment. Next, text-to-image generation pipeline synthesizes a large set of context images from the language descriptions. 
%
%


\noindent{\textbf{Context description images (CDI).}}
We assume that we have been given a small set of images that describes the context, which can be as small as 1 image. We call them {\em context description images (CDI)}. These images can come from an environment that contextually looks similar to the test environment. For example, if the test scenario includes a kitchen environment, the small set of kitchen images can be taken from any public dataset or from web images.



\noindent{\textbf{Image caption.}} Next step involves describing context information from the given CDIs. Language can be used to provide a concise description of the context. In addition, language is both compositional and explainable in nature. 
Given these advantages, we leverage those language descriptions to represent context. This raises a question: how can these language descriptions be generated? 

In order to automatically generate such a description of the context, we use image captioning methods. These captioning approaches generate a set of diverse textual captions for input images. 
Over the years, many image captioning methods have been developed \cite{rennie2017self,anderson2018bottom,gurari2020captioning}. We use self-critique sequence training (SCST) for image captioning work developed by Rennie et.al. \cite{rennie2017self}. 
%
%
SCST method have shown to achieve very good accuracy on different caption generation datasets including COCO caption generation challenge. However, it should be noted that our method does not rely on any specific image to caption generation method. Any popular or new SOTA method can be used in place of the SCST method.
For each CDI, $K$ language descriptions are generated using the SCST method. 
If there are $N$ CDIs, the caption generation steps provide $N \times K$ natural language descriptions.
Next, we use two approaches to create new context description sentences. 
Approach 1: Given small number of captions, we manually select context words (e.g., grass field), and then we fill them in a set of templates. For instance, a real image of grass filed. 
Approach 2: further, we also augment more related context words based on prior knowledge, for example, animals (dogs) can be found in forest and then fill them in a set of templates.
More over, it should be noted that the manual extraction of the context word could be substituted by noun extraction from sentences followed by interest class removal by using WordNet. We could also use ConceptNet to automatically provide augmented context words. 
Approach 2 can be used in extrame cases (zero-shot setting), where we have no CDI.




\noindent{\textbf{Image generation.}} Next step involves generating a diverse set of $M$ images for each language description of the context information. Recent time has seen a remarkable breakthrough in the text to image generation approach. 
Some popular frameworks for text to image data generation are DALL-E \cite{ramesh2021zero}, Stable Diffusion\cite{rombach2022high}, CogView \cite{ding2021cogview}, RU-DALLE \cite{sberbank31:rudalle}. They use large scale text-image data pairs to train large transformer models in a self-supervised fashion. This allows them to generate really high quality photorealistic images of any real world environment, incorporating semantics and compositional knowledge of the world. 
For each text description, we use DALL-E to generate $M$ images. This approach helps us to generate a total of $N \times K \times M$ images from $N$ CDIs. 
%
%
So, even for single context images, we are able to automatically generate a large set of new context images. 
As a post process, we use CLIP \cite{radford2021learning} to filter and further ensure that the generated images have no interest class.
%




\subsection{Language-driven Foreground generation} \label{sec:foreground_gen} 
We note that image generation process can also be used to generate diverse and high-quality foreground images.
We manually design several fixed prompt templates, such as \textit{A photo of \textless object\textgreater}, \textit{A realistic photo of \textless object\textgreater}, and \textit{A photo of \textless object\textgreater  in pure background}\footnote{We provide the full set of prompt templates in appendix.}. \textit{\textless object\textgreater} is replaced by various category labels and then the prompt is fed into DALL-E to generate high-quality iconic object foreground images. We highlight that our foreground generation is 0-shot, and only category labels are required.
Benefits of our process is that we can easily separate objects as engineering prompts allows to generate object on simple isolated backgrounds. 
We can then use generic unsupervised foreground extraction method \cite{qi2021open} to get the masks. 


The text-to-image generation model has several benefits. First, it is a compact version of web-scale image-text pair data. That makes it both portable and scalable. Also, being a generative model, the generation pipeline could create new scenarios that were not present in the training data. Furthermore, the synthetic nature of the data generation procedure allows our method to be privacy-preserving. 
Finally, language based data generation allows the model to be compositional. This whole approach allows us to generate large scale foreground object masks and background context images. Many of these generated images have been provided in the supplementary material.



\subsection{Compositional dataset generation} \label{cut and paste} 
We describe how to combine foreground object masks and background images obtained in \Cref{sec:syn_gen} and \Cref{sec:foreground_gen}. We composite foreground object masks onto the background images to create labelled synthetic data.

\noindent{\textbf{Label data generation.}} 
At each step a group of foreground object masks is selected and pasted into a sampled background image, and such procedure is repeated until all foreground object masks are pasted.
The foreground mask, after 2D geometric augmentation such as rotation and scaling, is pasted on a random location in the image. In addition, following \cite{dwibedi2017cut, ghiasi2021simple}, we apply a Gaussian blur on the object boundary with blur kernel as $\sigma$ to blend pasted foregrounds. 

\begin{figure*}[h]
\begin{center}
\includegraphics[width=\linewidth]{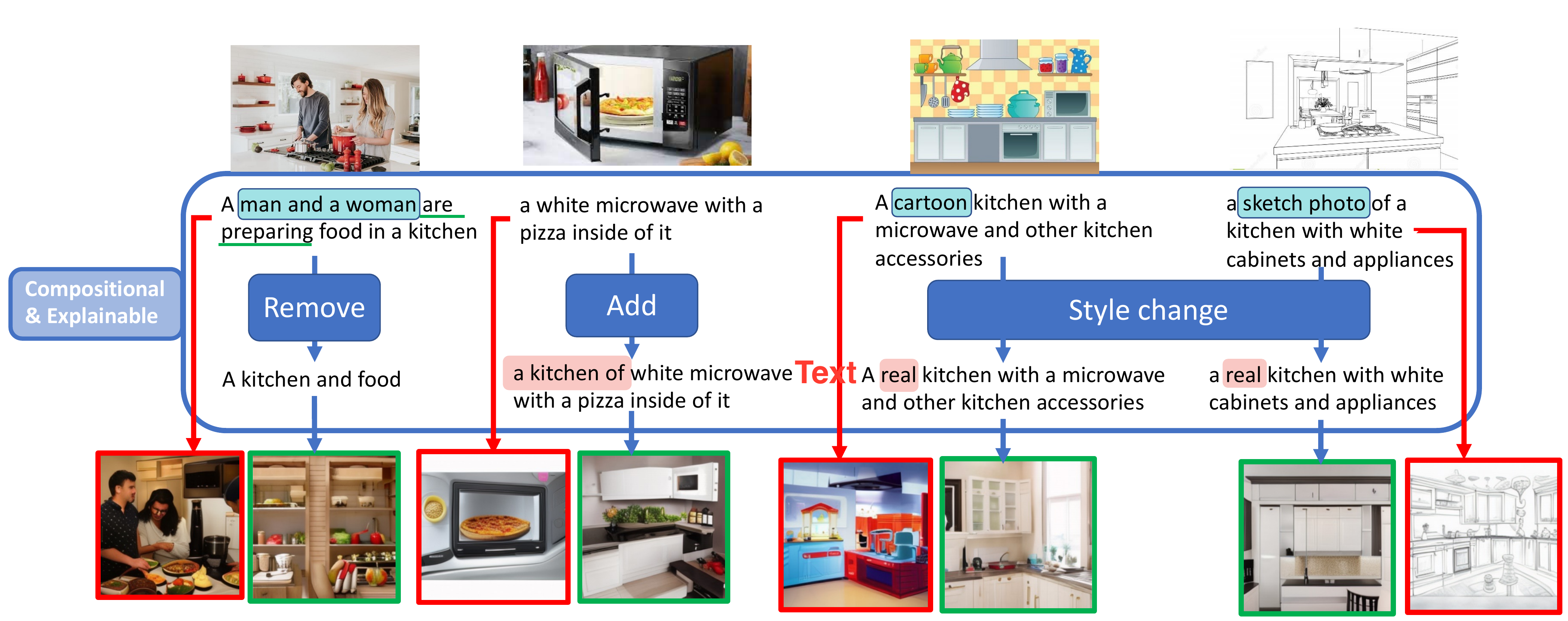}
\end{center}
\vspace{-5mm}
   \caption{ We highlight compositional and explainable properties of our method. Specifically, when the provide CDI can not perfectly describe the real test scenario, the compositional property of language can help to correct context description by remove/add/style change. For instance, if the initial description contains noisy information "man and a woman", we can directly intervene and remove the noise information to generate congruent context description. Note that, all the 4 example the test scenario are from GMU kitchen dataset. Images with \textcolor{red}{red} frame shows the generated image without language intervene and images with \textcolor{green}{green} frame shows the images after intervene.
   }
\label{fig:3}
\end{figure*}

\noindent{\textbf{Compositional data generation.}}
The natural language description allows compositional data generation. We can intervene in the language description to add or remove a key word (Fig.~\ref{fig:3}). For example, the word {\em kitchen} can be added to generate context. This allows us to generate new images with new context information. Similarly, we can remove some unwanted objects. For example, if initial context description involves people, we can remove people from generated images by simply not mentioning the word {\em{people}}. The text to image generation pipeline can then generate large set of images without people. More details are in the supplementary material.
\vspace{-1mm}
\section{Experiments}
\label{sec:experiment}
\vspace{-1mm}

We demonstrate the effectiveness of the proposed approach in automatically generating large datasets on three scenarios. First we evaluate our method on large scale object detection tasks on Pascal VOC and COCO datasets. 
Here, large scale training data has been created by synthetically generating the foreground masks for both VOC and COCO object classes, and background (context) images as discussed in the method section. 
Next, we highlight the benefits our approach on instance detection tasks. We consider GMU-Kitchen \cite{georgakis2016multiview}, Active Vision \cite{ammirato2017dataset}, and YCB-video datasets \cite{xiang2017posecnn}. We follow the training strategy of \cite{dwibedi2017cut} and since these datasets already provide object masks, we use our method to only generate good context images for downstream object instance detection tasks on the above three datasets. 
Finally, we also provide results highlighting compositional nature of our data generation process. For Pascal VOC and COCO dataset experiments, we use Stable Diffusion \cite{rombach2022high} as the text-to-image generation model, and for other experiments, we use RU-DALLE \cite{sberbank31:rudalle}.

\noindent{\textbf{Training procedure and evaluation criterion.}} We use faster RCNN \cite{ren2017faster} with ResNet-50 \cite{he2016deep} as the backbone to train both object detection and object instance detection network. Models have been trained till convergence for both the baselines and our approaches. In our experiments we set learning rate as $0.001$ with a weight decay $0.0005$. Furthermore, we report standard mean average precision (mAP) for the evaluation of object detection results.

\subsection{DALL-E for general large dataset: both foreground and context generation}

\begin{table*}[htpb]
    \centering
    \caption{DALL-E generated foreground and context images help solve object detection in PASCAL VOC \cite{everingham2010pascal} dataset. Column mAP is computed as average of IoU ranging from 50 to 95 with step size 5.}
    \label{tab:voc-new}
    \begin{tabular}{|l|l|c|c|c|cc|}
        \hline
        \textbf{EXP id} & \textbf{Dataset} & \textbf{\#CDI} & \textbf{Foreground} & \textbf{Background} & \textbf{mAP@50} & \textbf{mAP} \\
        \hline\hline
        G-1 EXP-1 & VOC 1.4k train & 1464 & \multicolumn{2}{c|}{Real} & 45.50 & 17.00 \\
        \hline\hline
        G-2 EXP-1 & \multirow{2}{*}{VOC 0 shot} & \multirow{2}{*}{0} & \multicolumn{2}{c|}{Syn} & \textbf{43.24} & \textbf{19.78} \\ \cline{1-1}\cline{4-7}
        G-2 EXP-2 &  &  & Syn & Web-bg & 38.35 & 17.76 \\
        \hline\hline
        G-3 EXP-1 & VOC 1 shot & \multirow{6}{*}{$20 \cdot 1$} & \multicolumn{2}{c|}{\multirow{2}{*}{Real}} & 0.14 & 0.04 \\ \cline{1-2}\cline{6-7}
        G-3 EXP-2 & \ \ + cut paste &  & \multicolumn{2}{c|}{} & 6.03 & 2.07 \\ \cline{1-2}\cline{4-7}
        G-3 EXP-3 & \ \ use syn fg &  & Syn + Real & Real & 37.97 & 17.53 \\ \cline{1-2}\cline{4-7}
        G-3 EXP-4 & \ \ only syn &  & \multicolumn{2}{c|}{Syn} & 44.24 & 20.63 \\ \cline{1-2}\cline{4-7}
        G-3 EXP-5 & \ \ syn + real &  & \multicolumn{2}{c|}{Syn + Real} & \textbf{45.62} & \textbf{21.45} \\ \cline{1-2}\cline{4-7}
        G-3 EXP-6 & \ \ COCO-bg + real &  & Syn + Real & COCO-bg + Real & 38.79 & 18.04 \\ \cline{1-2}\cline{4-7}
        \hline\hline
        G-4 EXP-1 & VOC 10 shot & \multirow{6}{*}{$20 \cdot 10$} & \multicolumn{2}{c|}{\multirow{2}{*}{Real}} & 9.12 & 2.35 \\ \cline{1-2}\cline{6-7}
        G-4 EXP-2 & \ \ + cut paste &  & \multicolumn{2}{c|}{} & 29.60 & 10.82 \\ \cline{1-2}\cline{4-7}
        G-4 EXP-3 & \ \ use syn fg &  & Syn + Real & Real & 48.14 & 21.62 \\ \cline{1-2}\cline{4-7}
        G-4 EXP-4 & \ \ only syn &  & \multicolumn{2}{c|}{Syn} & 44.59 & 20.72 \\ \cline{1-2}\cline{4-7}
        G-4 EXP-5 & \ \ syn + real &  & \multicolumn{2}{c|}{Syn + Real} & \textbf{51.82} &  \textbf{25.87} \\ \cline{1-2}\cline{4-7}
        G-4 EXP-6 & \ \ COCO-bg + real &  & Syn + Real & COCO-bg + Real & 48.20 & 23.95 \\ \cline{1-2}\cline{4-7}
        \hline
    \end{tabular}
\end{table*}

\begin{table*}[htpb]
    \centering
    \caption{DALL-E generated foreground and context images help solve object detection in COCO dataset}
    \label{tab:coco}
    \begin{tabular}{|l|l|c|c|c|cc|}
        \hline
        \textbf{EXP id} & \textbf{Dataset} & \textbf{\#CDI} & \textbf{Foreground} & \textbf{Background} & \textbf{mAP@50} & \textbf{mAP} \\
        \hline\hline
        EXP-1 & COCO 1 shot & \multirow{5}{*}{$80 \cdot 1$} & \multicolumn{2}{c|}{\multirow{2}{*}{Real}} & 1.47 & 0.92 \\ \cline{1-2}\cline{6-7}
        EXP-2 & \ \ + cut paste &  & \multicolumn{2}{c|}{} & 2.89 & 1.23 \\ \cline{1-2}\cline{4-7}
        EXP-3 & \ \ use syn fg &  & Syn + Real & Real & 17.87 & 8.64 \\ \cline{1-2}\cline{4-7}
        EXP-4 & \ \ only syn &  & \multicolumn{2}{c|}{Syn} & 16.22 & 7.66 \\ \cline{1-2}\cline{4-7}
        EXP-5 & \ \ syn + real &  & \multicolumn{2}{c|}{Syn + Real} & \textbf{20.82} & \textbf{10.63} \\ \cline{1-2}\cline{4-7}
        \hline
    \end{tabular}
\end{table*}
        

\subsubsection{PASCAL-VOC dataset}
\label{sec:voc}
We first evaluate our method on PASCAL VOC 2012 object detection task \cite{everingham2010pascal}. 

\noindent{\textbf{Dataset.}} The dataset has 20 foreground classes. The training and validation set consist of 1,464 images and 1,449 images, respectively with bounding box along with instance segmentation masks. We use the instance segmentation masks from the training set ground truth labels as our real foreground masks. 


\noindent{\textbf{Experiments set up.}}
As shown in Table.~\ref{tab:voc-new}, we conduct four groups of experiments. 
%
The first group (G-1) is the baseline experiment (EXP). We train the model using the whole 1,464 training images without any foreground or background synthesis.
The second group (G-2) is zero-shot setting, where we use no CDIs for background image generation (approach 1) but use only approach 2 method (class specific context words with prior knowledge) for context description (G-2 EXP-1). To compare our background context image generation pipeline with directly collecting background images from internet (web image), we build a web-image baseline (G-2 EXP-2) , where we collect same number of images using the same context description sentences with Google Search engine (EXP-2).

The thrid group (G-3) is VOC-1-shot, where we randomly sample 1 image per class (20 images total) from the real 1,464 training set as context description images (CDIs). The VOC-1-shot group consists of 6 EXPs.
%
(G-3 EXP-1) is the baseline which trains the object detection model on the 20 real training images without any augmentation of the images using cut-and-paste steps. 
(G-3 EXP-2) applies cut-and-paste method (\Cref{cut and paste}) into the baseline experiment. Specifically, we randomly paste the real foreground object masks from the CDIs onto the real training images.
For next EXPs, we first use information provided by the CDIs to generate context images from Stable Diffusion (Syn-back). We also use Stable Diffusion to synthesize foreground images and obtain foreground masks (Syn-obj) with our method.
For (G-3 EXP-3), we use both real foreground object mask and syn-obj as foregrounds and paste them on real CDIs as backgrounds.
%
(G-3 EXP-4) uses only synthetic foregrounds (syn-obj) and paste them on the synthetic backgrounds (syn-back).
(G-3 EXP-5) uses both real object mask and syn-obj as foreground, and also, for background, it use both real CDIs and syn-back. The combined real+syn foreground are then pasted on combined real+syn background. 
(G-3 EXP-6) substitutes the syn-back from EXP-5 with random COCO images (COCO-img), the rest setting is the same. We select COCO images that contains only the remaining 60 classes objects so that they are disjoint from VOC objects.

The fourth group (G-4) is VOC-10-shot that consists of similar 6 experiments as the 6 experiments in the third group. The only difference is the number of CDIs, where we randomly sample 10 image per class (200 images total) from the real 1,464 training set as context description images (CDIs). 
For all above EXP, we repeat the foregrounds and backgrounds accordingly so that the total number of synthesized dataset after paste is 60,000.

\noindent{\textbf{Results}}
Table.~\ref{tab:voc-new} shows the results of the four groups of different experiments.
First, simply using only synthetic foreground and background images in the zero-shot setting achieves 43.2\% on mAP@50 that is close to fully supervised setting that use all 1,464 Pascal training images (G-1 EXP-1). It should be noted that if we use web images (collected from Google search)(G-1 EXP-2), we observe a -4.89\% decrease on mAP@50 than our method. This shows that our proposed approach provides better context images over collected web images.

Next, we provide results when we have CDIs in VOC-1-shot (G-3) and VOC-10-shot (G-4) settings. 
In G-3 EXP-4, after adding 1-shot of CDIs, our approach with just synthetic foreground and synthetic background generated from Stable Diffusion achieves +38.21\% improvement over the baseline trained on 20 CDI images (1-shot per class) with cut-and-paste (G-3 EXP-2). 
Additionally, if we combine just around 1.4\% of real training images (20 CDIs' real foreground and real background) with our synthesized foreground and background, we observe better results (45.62 mAP@50 G-3 EXP-5) over baseline that use all 1,464 Pascal training images (G-1 EXP-1).
To further show the effectiveness of the synthesized background images, we substitute the Stable Diffusion synthesized background image in  G-3 EXP-5 with random COCO image (G-3 EXP-6), and we observer a -6.83\% decrease on mAP@50, which demonstrate that our used caption descriptions are important to generate congruent context images.

If we further increase the number of CDIs, where we use 10-shot real images per class (in total 200 images) (Group-2 EXPs), we observe significant improvement of our method over the corresponding baselines. For instance, combining only 200 real images with our synthetic pipeline (G-4 EXP-5), model trained with our dataset achieves 51.82\% on mAP@50 which has +6.32\% improvement over baseline that use all 1,464 Pascal training images (G-1 EXP-1). Note that the used real images (200 CDIs) in G-4 EXP-5 is just 13.7\% of all 1,464 training images used in G-1 EXP-1.

\vspace{-2mm}
\subsubsection{COCO dataset}
\label{sec:coco}

We also evaluate our method on COCO dataset \cite{lin2014microsoft}, we use the same experiment set up on model, training parameter, and evaluation metrics as the PASCAL VOC dataset in Sec.~\ref{sec:4.1}. We only conduct the 1-shot experiments. 

\noindent{\bf Results}
Table.~\ref{tab:coco} shows the results of different experiments in the one-shot setting.
First, our approach with just synthetic foreground and synthetic background generated from Stable Diffusion (EXP-4) achieves +14.75\% improvement over the baseline trained on 80 CDI images (1-shot per class) (EXP-1). 
In addition, we observe +13.33\% improvement over applying cut and paste augmentation on baseline methods (EXP-2).
Then, if we combine the 80 CDIs' real foreground and real background (CDIs) with our synthesized foreground and background, we observe further +4.60\% points improvement on mAP@50 with the help of our method (EXP-5) over (EXP-4).

\subsection{DALL-E for instance-specific dataset: only context generation}

Next we present results of our method on object instance detection tasks on three datasets: GMU-Kitchen, Active-Vision and YCB-video datasets. We follow the experiment setup of the prior approaches \cite{dwibedi2017cut} and use the object instance masks provided with the dataset. 

\noindent{\textbf{Dataset.}} GMU-kitchen \cite{georgakis2016multiview} consists of $6,728$ images from 9 kitchen scenes for 11 object instance classes. We use 3-fold cross-validation for testing on this data, following the object cut-and-paste paper \cite{dwibedi2017cut}. Active Vision consists of $4,000$ images with $2,000$ test images for 6 object classes.
We also evaluate our approach on the YCB-video objects \cite{xiang2017posecnn} that consists of 1000 training and 1000 test images for 21 object classes. Additional details about these datasets have been provided in the supplementary material including information about foreground masks for each dataset. 

\noindent{\textbf{Baselines.}} We evaluate our approach with other standard approaches for selecting context images. The main baseline consists of a comparison with the approach described in the Object Cut-and-Paste method \cite{dwibedi2017cut}. This approach involves selecting context images from the UW dataset \cite{georgakis2016multiview}. 
We also compare with other approaches: black background images, images from COCO dataset \cite{lin2014microsoft}, context description images (CDIs), random images generated by RU-DALLE. 
%
\vspace{-4mm}
\subsubsection{Results} \label{sec:GMU-Kitchen_sec}
\vspace{-2mm}
\label{sec:4.1}


\begin{figure}[t]
    \centering
    \includegraphics[width=\linewidth]{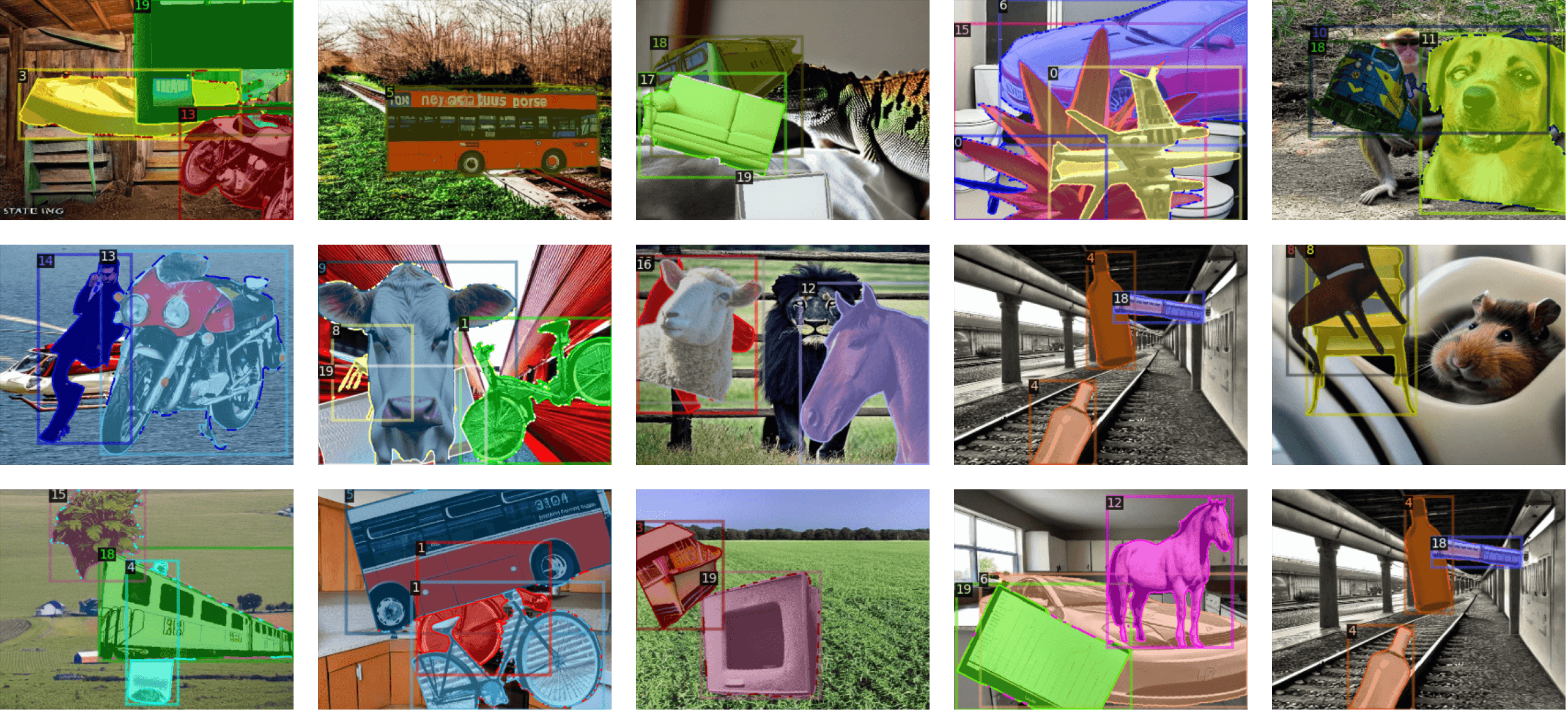}
    \caption{Training images generated by our pipeline: pasting foreground objects on DALL-E synthesized images with CDI form Pascal VOC dataset. }\label{fig:random_dalle_syn}
\end{figure}

\noindent{\textbf{GMU-Kitchen dataset.}} Quantitative results are shown in the Table \ref{table:gmu-kitchen-main}. We observe that our approach is able to get almost $2.1$ percentage points improvement over the Object Cut-and-Paste baseline. This highlights that our approach generates a better diverse set of context images compared to the images from the UW dataset. Furthermore, we also observe $37 \%$ points, $63.2\%$, and $16 \%$ points of improvements over black, CDIs, and COCO images respectively. 
%
Next, we conduct experiments to demonstrate that caption descriptions are important to generate congruent context images. To this end, we use random language descriptions to generate images from Ru-DALLE.
We observe almost $12\%$ points improvement of our approach over the random images from Ru-DALLE.
%
%
Finally, the use of real-world GMU training data combined with our synthetic data helps to achieve the best performance on the GMU test set, i.e., almost $5\%$ points improvement over training with real GMU training data. 
%
%
%
Compared to all the baselines, our approach is able to achieve better performance. This highlights the benefit of using our language-driven context image generation approach to synthesize congruent context images.

\begin{table}[t]
\begin{center}
\small
\begin{tabular}{c|c|c}
\hline
    Dataset & \#CDI  & mAP  \\
\hline
  Real GMU train& -  & 86.3   \\  
\hline  
  Black & 1500 & 41.2  \\
  CDI &  10 & 15.0  \\  
  Random (COCO) & 1500 & 62.8  \\ 
  Random (DALL-E) & 1500  & 66.8  \\  
  
  UW-Kitchen &  1500 & 76.1  \\
  
  DALL-E (ours) & 1500 & 78.3  \\ 
  
  DALL-E (ours) & 2400 & 80.1  \\
  \hline
  DALL-E (ours)+Real & 1500 & \textbf{91.4}  \\  
\hline
\end{tabular}
\end{center}
\vspace{-5mm}
\caption{Quantitative results on GMU-Kitchen dataset. We compare our approach with several prior approaches. Our approach achieve highest accuracy over the baselines. Combining GMU kitchen real training samples with our synthetic data yields the best results on this dataset. Real: real GMU training data as training set. \#CDI: the number of context description images. 
} 
\label{table:gmu-kitchen-main}
\end{table}




\begin{table}[b]
\begin{center}
\begin{tabular}{c|c|c}
\hline
    Dataset & Active Vision & YCB-video   \\
\hline
  UW-Kitchen    & 22.6  & 38.3   \\
  DALL-E (ours)    & \textbf{25.8}  &  \textbf{45.5} \\  
\hline
\end{tabular}
\end{center}
\vspace{-5mm}
\caption{Per class detection accuracy on Active Vision dataset. CC: Coca Cola, HB: honey bunches, HS: hunt's sauce, MR: mahatma rice, NV2: nature V2, RB: red bull. The baseline context is the real world UW-kitchen dataset.
} 
\label{table:active-vision-ycb}
\end{table}

\noindent{\textbf{Active Vision dataset.}} Quantitative results are shown in the Table \ref{table:active-vision-ycb}. We observe that our approach is able to get $3.2$ percentage points improvement over the object cut-and-paste baseline. This highlights that our approach generates better context images compared to the images from UW dataset. 




\noindent{\textbf{YCB-video dataset.}} Table \ref{table:active-vision-ycb} shows quantitative numbers of both the baseline and our approaches. Note that our approach can get $7.2\%$ points improvement over the Object Cut-and-Paste \cite{dwibedi2017cut} with UW-kitchen real context images. 
This highlights the benefit that our language-driven DALL-E generated images provide congruent context images compared to using real world images from other public datasets.

\vspace{-1.5mm}
\subsection{Compositional Model}
\label{sec:4.4}
\vspace{-1mm}
Here we demonstrate the compositional nature of our approach and highlight how language, as a self-interpretable modality with compositionality property, can provide several benefits for synthetic data generation. 


\begin{table}[t]
\begin{center}
\small
\begin{tabular}{c|c|c}
\hline
    Dataset  & No Intervention & After Intervention \\
\hline
 Cartoon Kitchen &  70.0 & \textbf{76.7} \\
 Skeleton Kitchen  &   64.6 & \textbf{74.8} \\  
 Objects in Kitchen &  71.8  & \textbf{77.0} \\
 Kitchens with Human  &   70.9 &  \textbf{76.9} \\  
\hline
\end{tabular}
\end{center}
\vspace{-5mm}
\caption{Quantitative results on GMU kitchen dataset highlighting the compositional benefits of our method in handling complex scenarios for example if CDIs come from out of distribution domians.
For example, for real world kitchen test environment (GMU kitchen), CDIs are cartoon kitchen images.
} 
\label{table:compositional}
\end{table}



\noindent{\textbf{Out of distribution CDIs.}} We first consider scenarios where the context description images are out-of-distribution images. For example, suppose the task is to do evaluation in the real kitchen environment, but the context description images are sketch or cartoon images of the kitchen. Even in these scenarios our approach can generate very good context images. This is achieved because the image caption method still works on these out-of-distribution images. Some of these out-of-distribution images, their corresponding captions and context images generated by our approach are provided in the  Fig.~\ref{fig:3}. 


\noindent{\textbf{Language intervention.}} 
Compositional nature of our language based context image generation allows us to {\em remove} unnecessary and noisy information or {\em add} relevant missing information from the original textual description of the CDIs. For instance, the test set scenario is a real  kitchen with people present in it. Language description of these CDIs may contain {\em people} as distractor that may hamper quality of the generated images and may effect the accuracy. 
Using our language-based image generation pipeline, we can remove the distractor by automatically detecting and removing the distractor word ({\em people}) from the caption by word detection before using them within DALL-E framework. Similar approach can also {\em add} relevant missing information to the original textual description of the CDIs.

We use the generated context images with the original caption from the noisy CDIs and the generated context images with modified captions to form two datasets and demonstrate the advantages of our compositional advantage in the Table \ref{table:compositional}. As seen, using context images after modification helps to improve performance by almost $6.7\%$ points, $10.2\%$ points, $5.3\%$ points, and $6\%$ points over generated images from non-modified caption on 4 out-of-distribution scenario (Cartoon kitchen, Skeleton kitchen, objects in Kitchen and Kitchen with human). 
Please refer to supplementary materials for more  qualitative images and further analysis. 

\vspace{-2mm}
\section{Conclusion}
\label{sec:conclusion}
\vspace{-2mm}

We have proposed a new paradigm to generate large scale labelled data for object detection and segmentation tasks using large vision and language-based text-to-image synthesis frameworks. We demonstrate effortless labelled data generation on popular benchmarks for object detection tasks. Computer vision models trained using these data improves the performance over the models trained with large real data. Thus reducing the need for expensive human labeling process. We also highlight the compositional nature of our data generation approach on out-of-distribution and zero-shot data generation scenarios. 

There are interesting extensions. Our approach can be used to solve other tasks like instance segmentation task, handling long-tail and imbalanced data problems. Next, the presented approach can be used to make models robust by iteratively generating the data where the detection model fails, which are then used to improve the robustness of the model. We believe the presented ideas of using vision-text models for automatically generating large labelled data opens door to democratizing computer vision models.

\noindent{\bf Acknowledgments} This work was supported in part by C-BRIC (one of six centers in JUMP, a
Semiconductor Research Corporation (SRC) program sponsored by DARPA),
DARPA (HR00112190134) and the Army Research Office (W911NF2020053). The
authors affirm that the views expressed herein are solely their own, and
do not represent the views of the United States government or any agency
thereof.


{\small
\bibliographystyle{ieee_fullname}
\bibliography{arxiv_main}
}

\clearpage
\section*{Appendix}

We provide details of our methodology in \Cref{sec:details_of_method}, including ablation studies of various components in our pipeline and some observations on those components' importance. Further we show that our pipeline can naturally be extended to other tasks such as low resource instance segmentation in \Cref{sec:instance_seg}. 
Moreover, we demonstrate that blending real data can further boost the performance in \Cref{sec:blend_real_for_improvement}. 
Lastly, in \Cref{sec:method_enables_compositional} we show that by utilizing compositionality of natural language we can narrow the domain gap and tackle out-of-domain problem.
For completeness, we give model prediction in \Cref{sec:model_prediction}.

\section{Details of our method}\label{sec:details_of_method}
We provide more details of our synthesized dataset here. Consider VOC \cite{everingham2010pascal} dataset with 20 objects\footnote{Synthetic dataset for other datasets like COCO \cite{lin2014microsoft} is generated similarly.}. 


For context background generation, approach 1 uses CDI to provide context description. Specifically,
$K=2$ captions are generated for each of the CDI, and each caption generates $M=80$ images. 
As a post-processing step, we use CLIP \cite{radford2021learning} to filter $M$ images such that only $30$ context images remain. 
Therefore for each CDI we generate $K \cdot M = 2 \cdot 30 = 60$ images.  
In 10 shot we have a total of $20 \cdot 10 \cdot 60 = 12,000$ synthesized context images; while for 1 shot we have a total of $20 \cdot 1 \cdot 60 = 1,200$ context images. 
For approach 2, we design 16 templates, and 600 images are generated for each template. 
We observe that generated images are of high quality and generally do not contain interested objects.
Therefore we prune 5\% images by CLIP only, which results in 9,120 synthetic backgrounds.
%

For synthesizing images for the foreground objects (\Cref{sub:foreground_crucial}), we generate 500 images for each of the $6$ templates and also use CLIP \cite{zhang2020learning} to select 250 best images. Thus we have a total of $20 \cdot 6 \cdot 250 = 30,000$ synthesized foreground images for the 20 objects in VOC.
%
Since foreground mask extraction process will drop some bad examples by design, the number of final processed extracted foregrounds is $26,685$. 
We repeat cut-and-paste such that the final pasted dataset have size $60,000$. Specifically, in each step, we first randomly select a background image, then 4 randomly selected foreground masks are pasted at random locations in the background image. Note that our generated dataset contains no duplicate.

Fig.~\ref{fig:final-composition} shows more example training images generated by our pipeline on PASCAL VOC dataset: both foreground object and background context images are generated by our method with Stable Diffusion.

\begin{table*}[h]
    \centering
    \begin{tabular}{ccc}
    \hline
        A photo of \textless object \textgreater & A realistic photo of \textless object \textgreater & A photo of \textless object \textgreater in pure background \\
        \hline
        \textless object \textgreater in a white background &  \textless object \textgreater without background & \textless object \textgreater isolated on white background \\
        \hline
    \end{tabular}
    \caption{Six manually designed templates for generating foreground images. Here \textless object \textgreater will be replaced by label names such as ``bus''. The design philosophy is to put object in a clean background for easy foreground extraction.}
    \label{tab:fg_template}
\end{table*}

\subsection{CLIP as variance reduction in quality control}\label{sub:CLIP_is_beneficial}

As shown in the main paper Fig. 2, to control the quality and correctness of synthesized image with Stable Diffusion, we use CLIP \cite{radford2021learning} to filter and rank the quality of Stable Diffusion synthesized context background images. This ensures that the generated background images have no interest classes. Specifically, CLIP ranks images with two rules: 1) images are semantically similar to the input caption to Stable Diffusion, 2) images are semantically dissimilar to any of the interest class labels that we explicitly detect and substituted in the previous steps. For each of the $M$ images generated for each caption of every CDI we only keep 30 images.



\begin{table*}[htpb]
    \centering
    \begin{tabular}{|l|l|c|c|c|ccc|}
        \hline
        \textbf{EXP id} & \textbf{Dataset} & \textbf{\#CDI} & \textbf{Foreground} & \textbf{Background} & \textbf{mAP@50} & \textbf{mAP@75} & \textbf{mAP} \\ 
        \hline\hline
        G-4 EXP-3 & \ \ use syn fg & \multirow{6}{*}{$20\cdot 10$} & Syn + Real & Real & 48.14 & 16.15 & 21.62 \\ \cline{1-2}\cline{4-8}
        \textcolor{blue}{G-4 EXP-3-50\%} & \ \ use 50\% syn fg &  & Syn (50\%) + Real & Real & 29.17 & 10.10 & 13.06 \\ \cline{1-2}\cline{4-8}
        \textcolor{olive}{G-4 EXP-3-30\%} & \ \ use 30\% syn fg &  & Syn (30\%) + Real & Real & 23.90 & 2.90 & 7.67 \\ \cline{1-2}\cline{4-8}
        G-4 EXP-5 & \ \ syn + real &  & \multicolumn{2}{c|}{Syn + Real} & \textbf{51.82} & \textbf{22.58} & \textbf{25.87} \\ \cline{1-2}\cline{4-8}
        \textcolor{blue}{G-4 EXP-5-50\%} & \ \ syn (50\%) + real &  & \multicolumn{2}{c|}{Syn (50\%) + Real} & 45.96 & 16.81 & 21.32 \\ \cline{1-2}\cline{4-8}
        \textcolor{olive}{G-4 EXP-5-30\%} & \ \ syn (30\%) + real &  & \multicolumn{2}{c|}{Syn (30\%) + Real} & 44.43 & 14.04  & 19.25 \\ \cline{1-2}\cline{4-8}
        \hline
    \end{tabular}
    \caption{Ablation study on synthetic foreground in PASCAL VOC \cite{everingham2010pascal} dataset. We observe an increase in performance when the number of synthetic foregrounds increases. \textcolor{blue}{EXP-*-50\%} experiment uses 50\% of the synthetic foregrounds while \textcolor{olive}{EXP-*-30\%} experiment uses 30\%.}
\label{tab:voc-ablation-fg}
\end{table*}
\subsection{High quality foreground masks are crucial} \label{sub:foreground_crucial}
We observe Stable Diffusion is able to synthesize high quality foregrounds as well. Given six manually designed templates provided in \Cref{tab:fg_template}, we replace each \textless object \textgreater with desired class labels, e.g. ``Bus isolated on white background'', and our assumption is that by the design of our templates Stable Diffusion will generate easy-to-recognize (e.g. centered and clean) target foregrounds in a background that is easy to differentiate between foreground. Fig.~\ref{fig:fg-extract} first row shows the generated image with foreground objects.

\noindent{\bf Foreground object mask extraction}
After high quality foreground images with easy-to-separate background are obtained, in order to retrieve the foreground object masks, we use off-the-shelf unsupervised segmentation methods such as entity segmentation \cite{qi2021open} to produce image segments. With these images of known categories, we can train an image classifier effortlessly. Then we use the image classifier to select the foreground segment. Fig.~\ref{fig:fg-extract} shows the extracted foreground masks.

\begin{figure*}[h]
\begin{center}
\includegraphics[width=\linewidth]{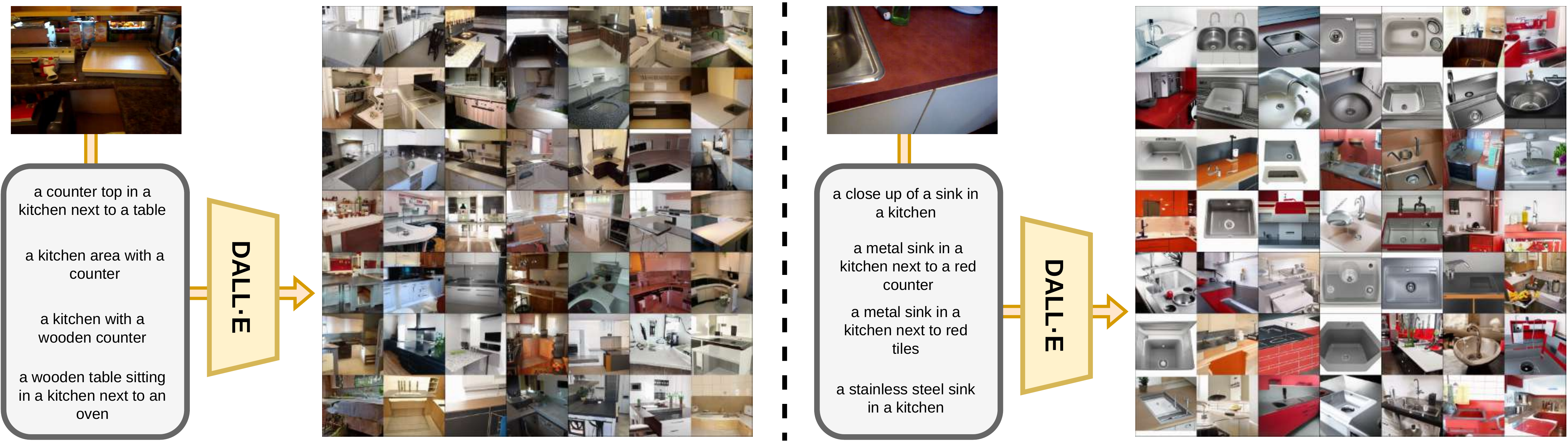}
\end{center}
   \caption{Context-images generated from our pipeline. We note that generated images are coherent to input captions.}
\label{fig:syn_example_appdenix}
\end{figure*}

\noindent{\bf The more foreground object masks, the better}
We note that our foreground generation is zero-shot, and only class labels are required. We can potentially generate as many foreground masks as we want, and we observe performance improvement as the number of foreground masks increase. In \Cref{tab:voc-ablation-fg}, we showed that using 
\textcolor{blue}{50\% of synthetic foregrounds} outperforms
\textcolor{olive}{30\% of foregrounds} in both settings (EXP-3 and EXP-5) and in 10 shot setting; while 100\% is even better than \textcolor{blue}{50\%}. 
We emphasize that since this approach is zero-shot and unsupervised, this is \textit{almost a free lunch} and we can perhaps further improve performance by increasing the number of foregrounds, which we leave to future work. 
Our method can be viewed as \textit{extracting information from generative model to enhance discriminative model}.

\begin{table*}[htpb]
    \centering
    \begin{tabular}{|l|l|c|c|c|ccc|}
        \hline
        \textbf{EXP id} & \textbf{Dataset} & \textbf{\#CDI} & \textbf{Foreground} & \textbf{Background} & \textbf{mAP@50} & \textbf{mAP@75} & \textbf{mAP} \\ 
        \hline\hline
        G-3 EXP-1 & VOC 1 shot & \multirow{3}{*}{$20 \cdot 1$} & \multicolumn{2}{c|}{Real} & 0.00 & 0.00 & 0.00 \\ \cline{1-2}\cline{4-8}
        G-3 EXP-3 & \ \ use syn fg &  & Syn + Real & Real & 42.23 & 20.62 & 21.80 \\ \cline{1-2}\cline{4-8}
        G-3 EXP-5 & \ \ syn + real &  & \multicolumn{2}{c|}{Syn + Real} & \textbf{46.74} & \textbf{24.03} & \textbf{24.81} \\
        \hline
        G-4 EXP-1 & VOC 10 shot & \multirow{3}{*}{$20 \cdot 10$} & \multicolumn{2}{c|}{Real} & 24.50 & 4.96 & 9.24 \\ \cline{1-2}\cline{4-8}
        G-4 EXP-3 & \ \ use syn fg &  & Syn + Real & Real & 51.29 & 29.15 & 29.11 \\ \cline{1-2}\cline{4-8}
        G-4 EXP-5 & \ \ syn + real &  & \multicolumn{2}{c|}{Syn + Real} & \textbf{55.19} & \textbf{30.58} & \textbf{30.77} \\ \cline{1-2}\cline{4-8}
        \hline
    \end{tabular}
    \caption{Instance segmentation results on PASCAL VOC \cite{everingham2010pascal} dataset. We demonstrate that our pipeline can be extended to other tasks and can outperform direct supervision in low resource setting by a large margin. Our results are competitive even in 1 shot.}
\label{tab:instance_seg}
\end{table*}

\section{Our method enhances low resource instance segmentation}\label{sec:instance_seg}
Although we mainly discuss object detection task in the main paper, we highlight that our method is not restricted to detection tasks only. In this section we demonstrate the effectiveness of our pipeline in instance segmentation. This is a simple extension since during the paste process we obtain not only bounding box but also segmentation task.

We report the performance in \Cref{tab:instance_seg}. Similar to experiments we layout in the main paper, we apply our method in low resource setting on PASCAL VOC \cite{everingham2010pascal} dataset. We conduct experiments on 10 shot and 1 shot. We observe a strong performance in both low resource settings, in which our method significantly outperforms the direct supervision (+30.69\% improvement on mAP@50 in 10-shot setting). In 1 shot scenario in which direct supervision can not make any correct prediction, our method can achieve 46.74 mAP@50. 




\begin{table*}[!]
\begin{center}
\begin{tabular}{c|c|c|c|c|c|c|c|c|c|c|c|c}
\hline
    Dataset  & CC & CM & HB & HS & MR & NV1 & NV2 & PO & PS & Pbbq & RB & mAP  \\
\hline
  DALL-E (ours)  &  79.0  & 92.9 & 90.4 & 44.9 & 77.0 & 92.1 & 88.0 & 77.5 & 64.1 & 75.7  &  80.2 & 78.3  \\
  100\% Real  &  81.9  & 95.3 & 92.0 & 87.3 & 86.5 & \textbf{96.8} & 88.9 & \textbf{80.5} & 92.3 & 88.9  &  58.6 & 86.3  \\
  \hline
  DALL-E (ours) + 10\% Real  & 90.5 & 96.9 & 93.2 & 74.0 & 60.4 & 90.7 & 86.5 & 48.7 & 97.7 & 86.4 & 72.1 & 81.6 \\
  DALL-E (ours) + 40\% Real  & 91.8 & 97.4 & 94.5 & 84.9 & 75.1 & 90.7 & 78.6 & 52.1 & 96.9 & 87.6 & 77.9 & 84.3 \\
  DALL-E (ours) + 70\% Real  & 92.7 & 98.2 & 95.2 & \textbf{90.9} & 88.0 & 93.1 & 89.7 & 50.3 & 97.6 & 92.2 & 78.3 & 87.9 \\
  DALL-E (ours) + 100\% Real  & \textbf{94.4} & \textbf{98.2} & \textbf{95.2} & 90.7 & \textbf{92.5} & 94.1 & \textbf{93.0} & 72.8 & \textbf{98.3} & \textbf{98.7} & \textbf{79.8} & \textbf{91.4}  \\  
\hline
\end{tabular}
\end{center}
\caption{We highlight that our synthesized data together with 70 $\%$ amount of real data achieves better performance than full (100 $\%$) set of real data only. This highlights the benefit of our approach in reducing total human efforts. DALL-E (ours) means DALL-E synthesized 1500 diverse images (use UW as CDI). Top row terms are: CC: Coca Cola, CM: Coffee mate, HB: honey bunches, HS: hunt's sauce, MR: mahatma rice, NV1: nature V1, NV2: nature V2, PO: palmolive orange, PS: pop secret, Pbbq: pringles bbg, RB: red bull.}
\label{table:vary_real}
\end{table*}

\section{Blending real data for further improvement}\label{sec:blend_real_for_improvement}

We first demonstrate the effect of incorporating different percentages of real-world training images together with our synthesized images for training object detection models.
We conduct experiments on the GMU kitchen \cite{georgakis2016multiview} test set and the real-world training images are from the GMU kitchen training dataset (100\% set contains 3837 images). 
This experimental setup is similar to the one followed in the Object cut-and-paste paper \cite{dwibedi2017cut}.
All the results are provided in the 
Table \ref{table:vary_real}.
%
%
%
We highlight the mAP accuracy of training with all synthetic data plus 10\%, 40\%, 70\% and 100\% real data.

Observe how using only a subset of real-world data ($70 \%$) with our synthesized images achieves better performance than full (100\%) real-world data only. This suggests the advantages of our data generation approach saving the amount of human efforts required in labeling the real-world data significantly.
Further, we also observe that accuracy gradually improves from $78.3 \%$ to $91.4 \%$ as we increase the amount of real-world data.
%






\begin{table*}
\begin{center}
\begin{tabular}{c|c|c|c|c|c|c|c|c|c|c|c|c|c}
\hline
    Dataset & \#CDI & CC & CM & HB & HS & MR & NV1 & NV2 & PO & PS & Pbbq & RB & mAP  \\
\hline
  Real GMU train& - & 81.9 & 95.3 & 92.0 & 87.3 & 86.5 & 96.8 & 88.9 & 80.5 & 92.3 & 88.9 & 58.6 & 86.3   \\  
\hline  
  Black & 1500 & 42.3  & 62.4 & 64.7 & 5.3 & 23.3 & 61.1 & 56.5 & 75.3 &  1.6 & 26.7 & 33.9  & 41.2  \\
  CDI &  10 & 51.4 & 26.4 & 2.1 & 12.2 & 12.1 & 0.4 & 0.1 & 1.0 & 0.1 & 29.8 & 30.0 & 15.0  \\  
  Random (COCO) & 1500 & 50.7 & 80.1 & 77.5 & 15.3 & 32.2 & 81.7  & 87.9 & 71.7 & 66.8 & 59.0 & 68.5 & 62.8  \\ 
  Random (DALL-E) & 1500 & 64.8 & 86.9 & 78.7 & 49.2 & 62.2 & 84.8  & 83.8  & 72.6 & 70.9 & 57.2 & 24.1  & 66.8  \\  
  
  
  UW-Kitchen &  1500  & 75.7  & 91.1 &  87.7 & 51.6 & 66.5 & 91.5 & 88.7  & 76.2 & 63.2 & 70.5 & 75.2 &  76.1  \\
  
  DALL-E (ours) & 1500 & 79.0 & 92.9 & 90.4 & 44.9 & 77.0 & 92.1 & 88.0 & \textbf{77.5} & 64.1 & 75.7 & 80.2 & 78.3  \\ 
  
  DALL-E (ours) & 2400 &  79.5  & 93.4 & 88.5 & 59.0 & 71.5 & 91.4 & 88.1 & 76.1 & 78.7 & 75.7  &  \textbf{80.6} & 80.1  \\
  \hline
  DALL-E (ours)+Real & 1500 & \textbf{94.4} & \textbf{98.2} & \textbf{95.2} & \textbf{90.7} & \textbf{92.5} & \textbf{94.1} & \textbf{93.0} & 72.8 & \textbf{98.3} & \textbf{98.7} & 79.8 & \textbf{91.4}  \\  
\hline
\end{tabular}
\end{center}
\caption{Quantitative results on GMU-Kitchen dataset. We compare our approach with several prior approaches. Our approach achieves highest accuracy over the baselines. Combining GMU kitchen real training samples with our synthetic data yields the best results on this dataset. Real: real GMU training data as training set. \#CDI: the number of context description images. 
} 
\label{table:gmu-kitchen}
\end{table*}




\section{Our method enables compositional manipulation}\label{sec:method_enables_compositional}

\noindent{\bf Context image generation from only one CDI}

Here we provide some visualization examples of context images generated from just \textit{one} given Context Description Images (CDI). To demonstrate that our model is very generic and can generate a large set of diverse context images from given input as little as one image, we include some examples of generated images from our pipeline as shown in Fig. ~\ref{fig:syn_example_appdenix}. 


\noindent{\bf  Compositional property of language can help to correct context description by remove/add/style change.}

In this section, we demonstrate more results for the compositional experiments present in section 4.3 of the main paper.

In table 6 of the main paper, we evaluate the models with synthetic data generated before intervention and after an intervention. Here in table~\ref{table:compositional_appendix}, we provide additional results of training models just on the CDIs.
%
We observe that the model can not yield good results due to the insufficient amount of training instances and the large domain gap. However, if we apply our method without intervention, we can get a significant performance boost by providing diverse training instances. Furthermore, by applying intervention, we can narrow the domain gap and yield even more performance gain, reinforcing the effectiveness of our approach. 
For instance,  Fig.\ref{fig:cartoon_appendix} visualize the context images generated by DALL-E for noisy and out-of-distribution CDIs. In the left column, inputs CDIs are
cartoon kitchen which provide noise information about environment. The compositional property allow us change the style from cartoon
to real and generate high-quality context images (right column)



Here we show qualitative examples of out-of-distribution images with their corresponding captions and how the compositional property of our method can help to correct context descriptions by remove/add and style change.  

Specifically, we include figures that exemplify our generated images in the compositional experiments. Fig.~\ref{fig:cartoon_appendix}, fig.~\ref{fig:sketch_appendix}, fig.~\ref{fig:object_appendix}, and fig.~\ref{fig:human_appendix} are generated results before and after intervention for the experiment \texttt{Cartoon Kitchen}, \texttt{Skeleton Kitchen}, \texttt{Objects in Kitchen}, and \texttt{Kitchens with Human} in the Table.~\ref{table:compositional_appendix}, respectively.

\begin{table}[h]
\begin{center}
\footnotesize
\scalebox{0.9}{
\begin{tabular}{c|c|c|c}
\hline
    Dataset  & only CDI & No Intervention & After Intervention \\
\hline
 Cartoon Kitchen & 11.2 & 70.0 & \textbf{76.7} \\
 Skeleton Kitchen  & 10.3 & 64.6 & \textbf{74.8} \\  
 Objects in Kitchen & 9.4 & 71.8  & \textbf{77.0} \\
 Kitchens with Human  & 10.2 & 70.9 &  \textbf{76.9} \\  
\hline
\end{tabular}
}
\end{center}

\caption{Quantitative results on GMU kitchen dataset highlighting the compositional benefits of our method in handling complex scenarios for example if CDIs come from out-of-distribution domains.
For example, for real-world kitchen test environments (GMU kitchen), CDIs are cartoon kitchen images.
} 
\label{table:compositional_appendix}
\end{table}



\begin{figure*}[h]
\begin{center}
\includegraphics[width=\linewidth]{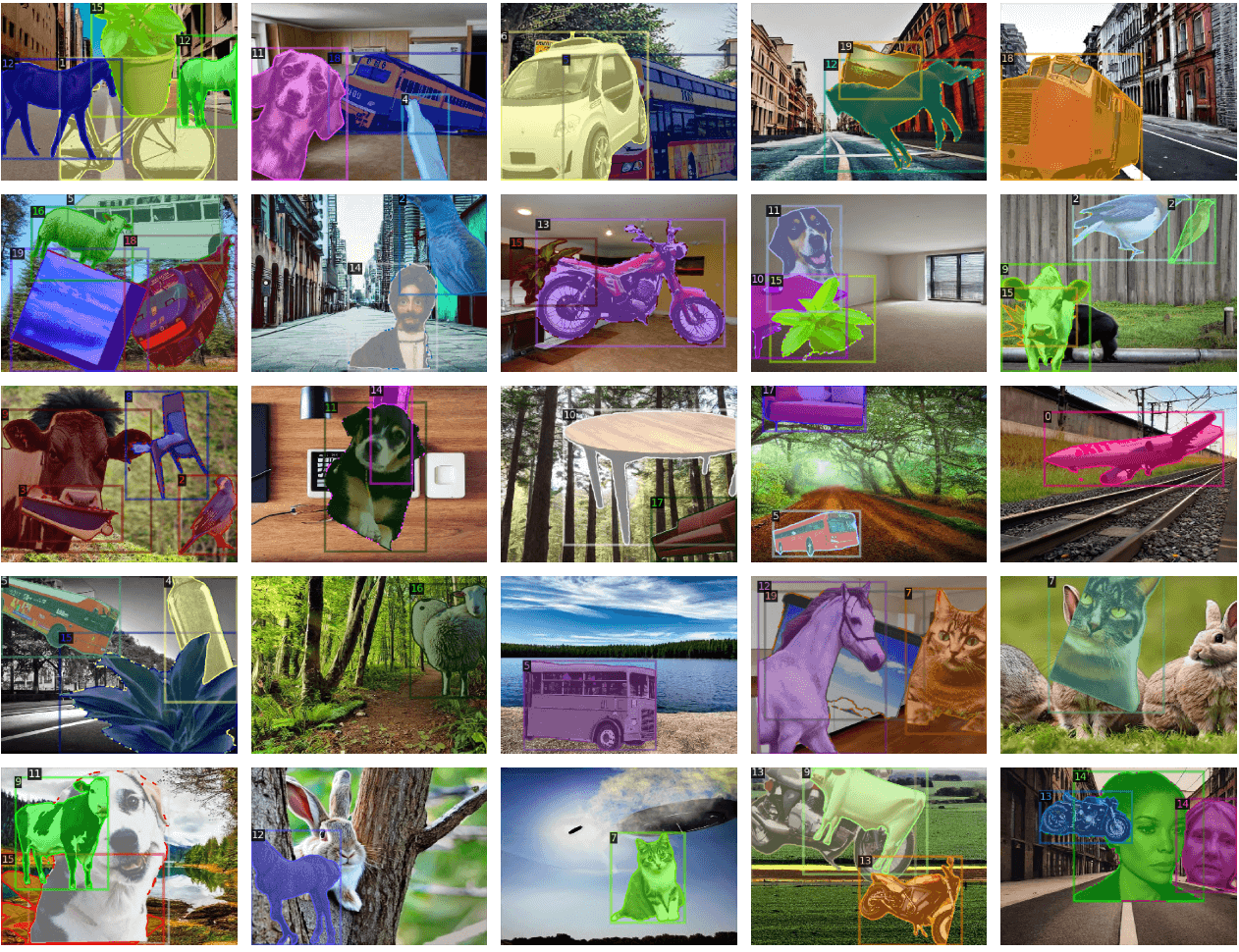}
\end{center}
   \caption{Training images generated by our pipeline for PASCAL VOC dataset: both foreground object and background context images are generated by our method with Stable Diffusion. }
\label{fig:final-composition}
\end{figure*}

\begin{figure*}[h]
\begin{center}
\includegraphics[width=0.95\linewidth]{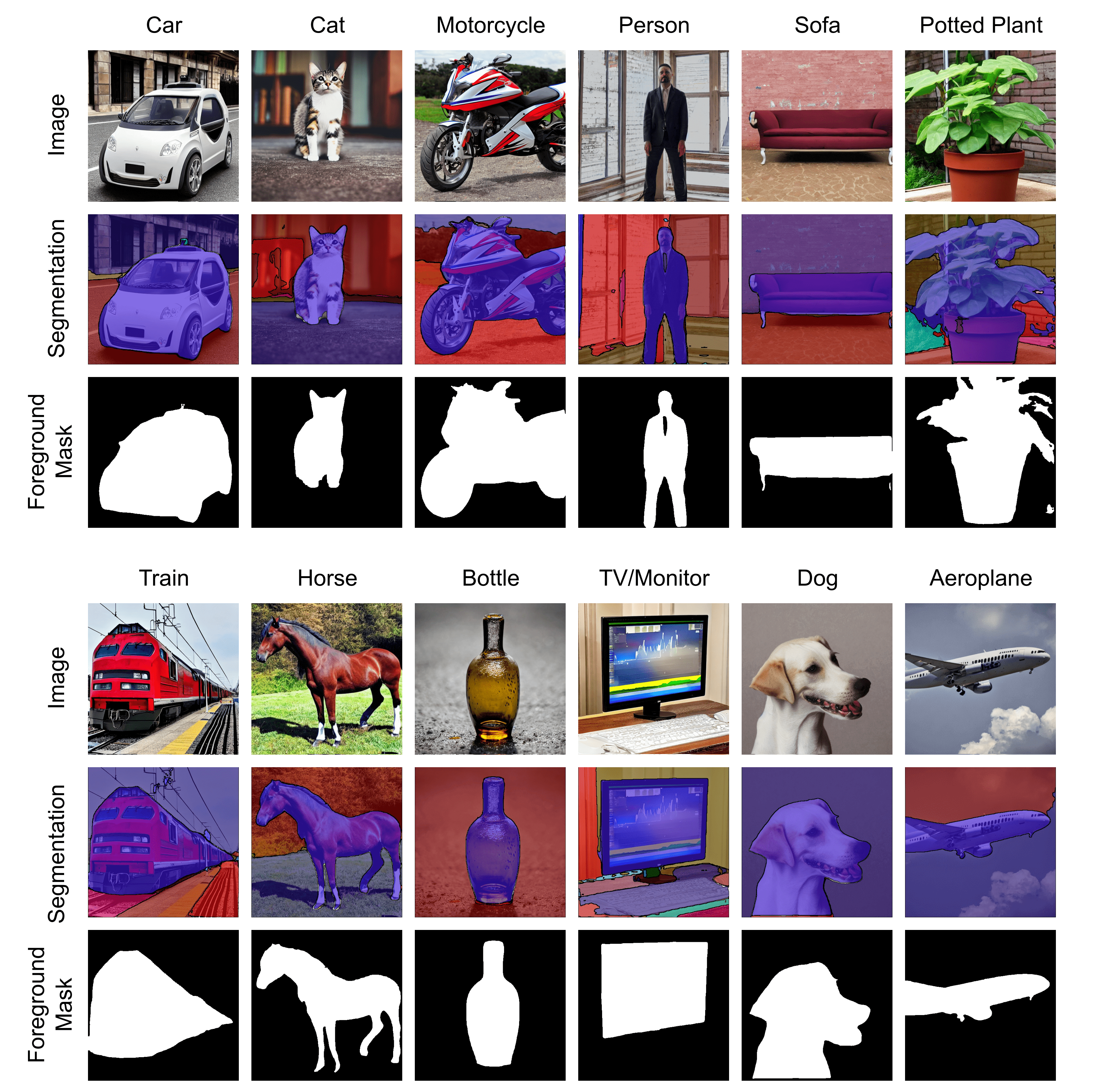}
\end{center}
   \caption{Examples of foreground segment extraction from generated images. }
\label{fig:fg-extract}
\end{figure*}

\begin{figure*}[h]
\begin{center}
\includegraphics[width=0.8\linewidth]{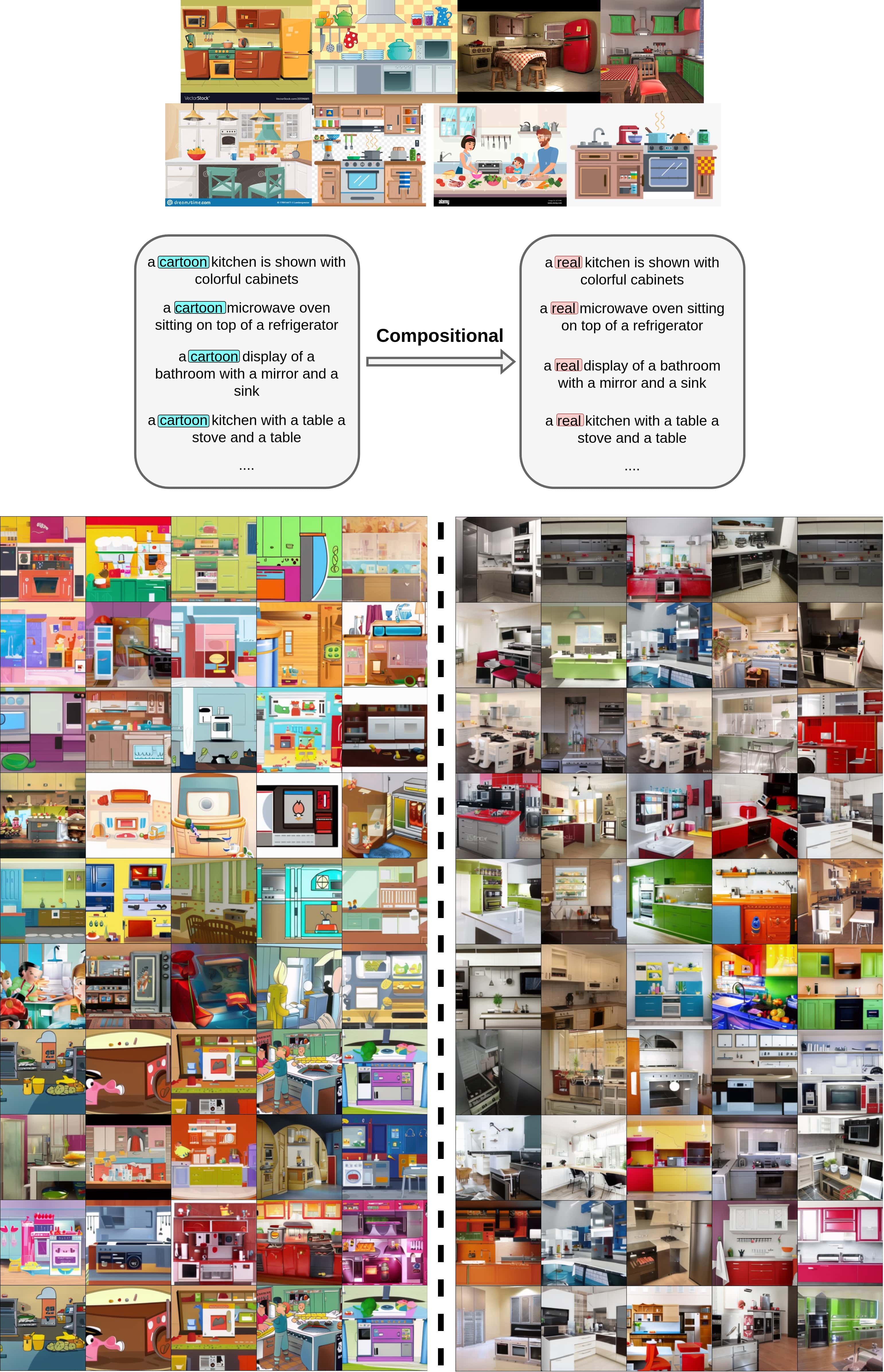}
\end{center}
   \caption{Visualization of context images generated by DALL-E for noisy and out-of-distribution CDIs. In the left column, inputs CDIs are cartoon kitchen which provide noise information about environment. The compositional property allow us \textbf{change the style} from cartoon to real and generate high-quality context images (right column).}
\label{fig:cartoon_appendix}
\end{figure*}

\begin{figure*}[h]
\begin{center}
\includegraphics[width=0.8\linewidth]{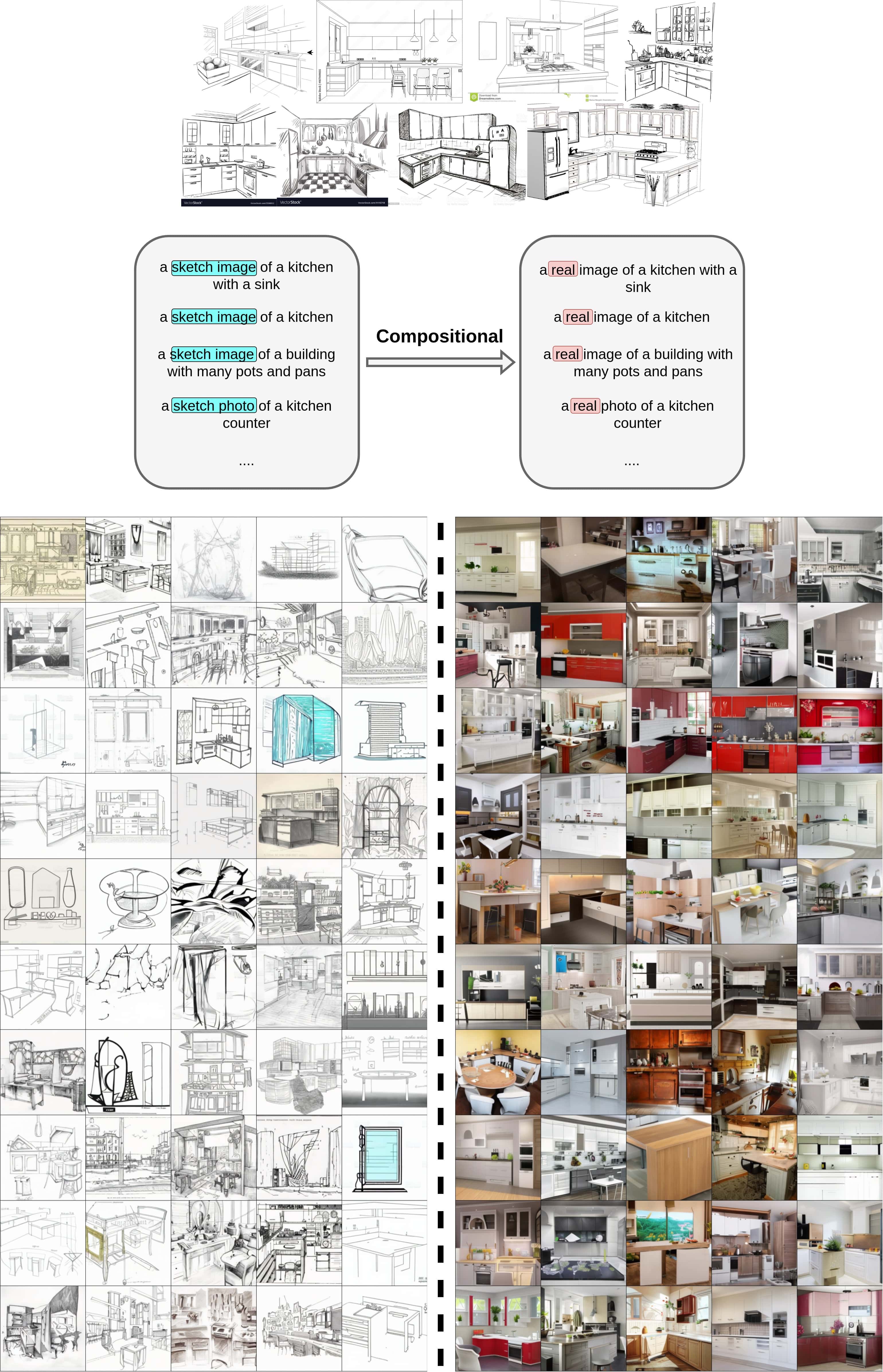}
\end{center}
   \caption{Visualization of context images generated by DALL-E for noisy and out-of-distribution CDIs. In the left column, inputs CDIs are sketch kitchen which provide noise information about environment. The compositional property allow us \textbf{change the style} from sketch to real and generate high-quality context images (right column).}
\label{fig:sketch_appendix}
\end{figure*}

\begin{figure*}[h]
\begin{center}
\includegraphics[width=0.8\linewidth]{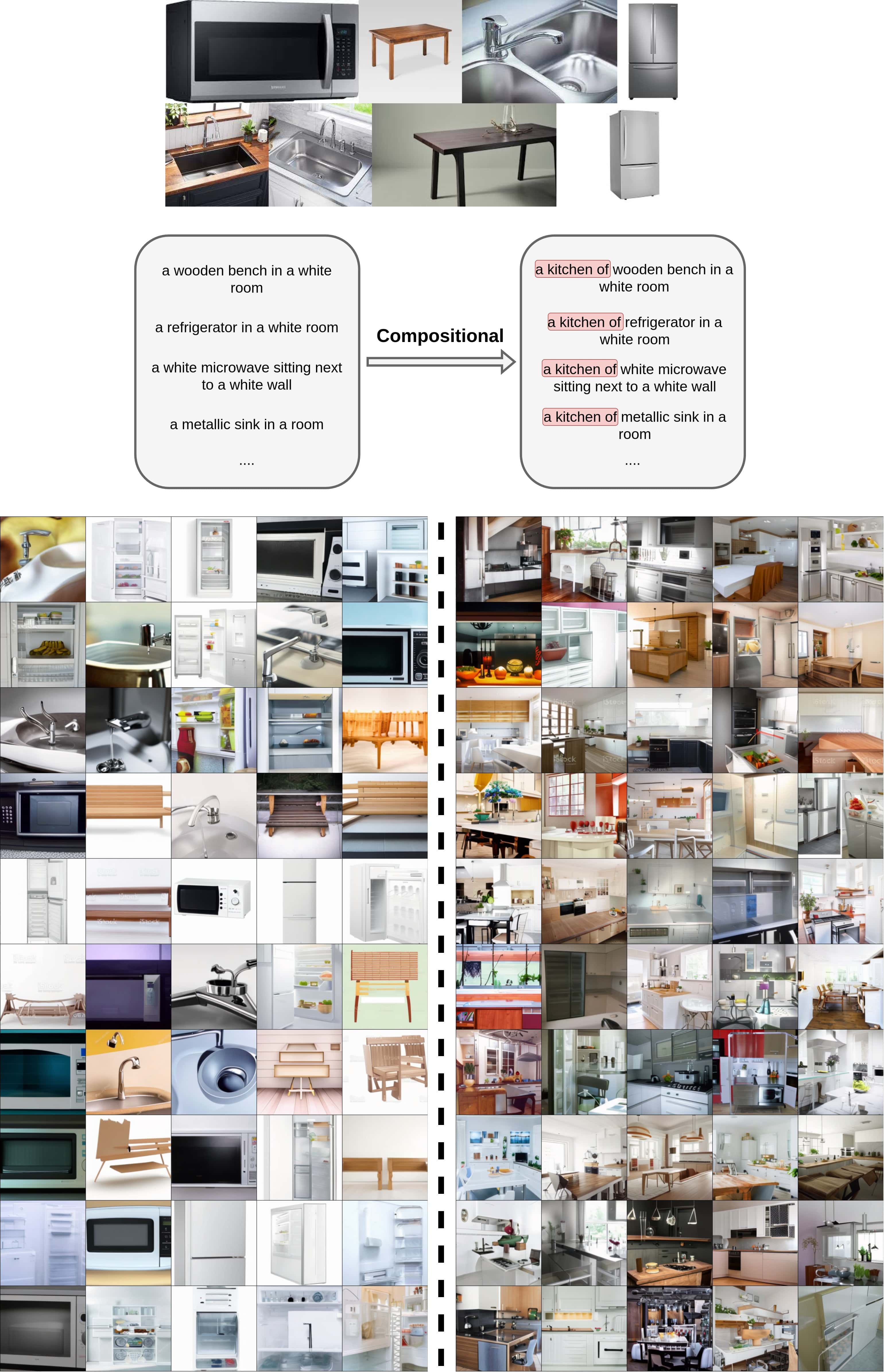}
\end{center}
   \caption{Visualization of context images generated by DALL-E for noisy and out-of-distribution CDIs. In the left column, inputs CDIs do not convey full knowledge about the context. The compositional property allow us \textbf{add} the context words to captions and generate high-quality context images (right column).}
\label{fig:object_appendix}
\end{figure*}

\begin{figure*}[h]
\begin{center}
\includegraphics[width=0.8\linewidth]{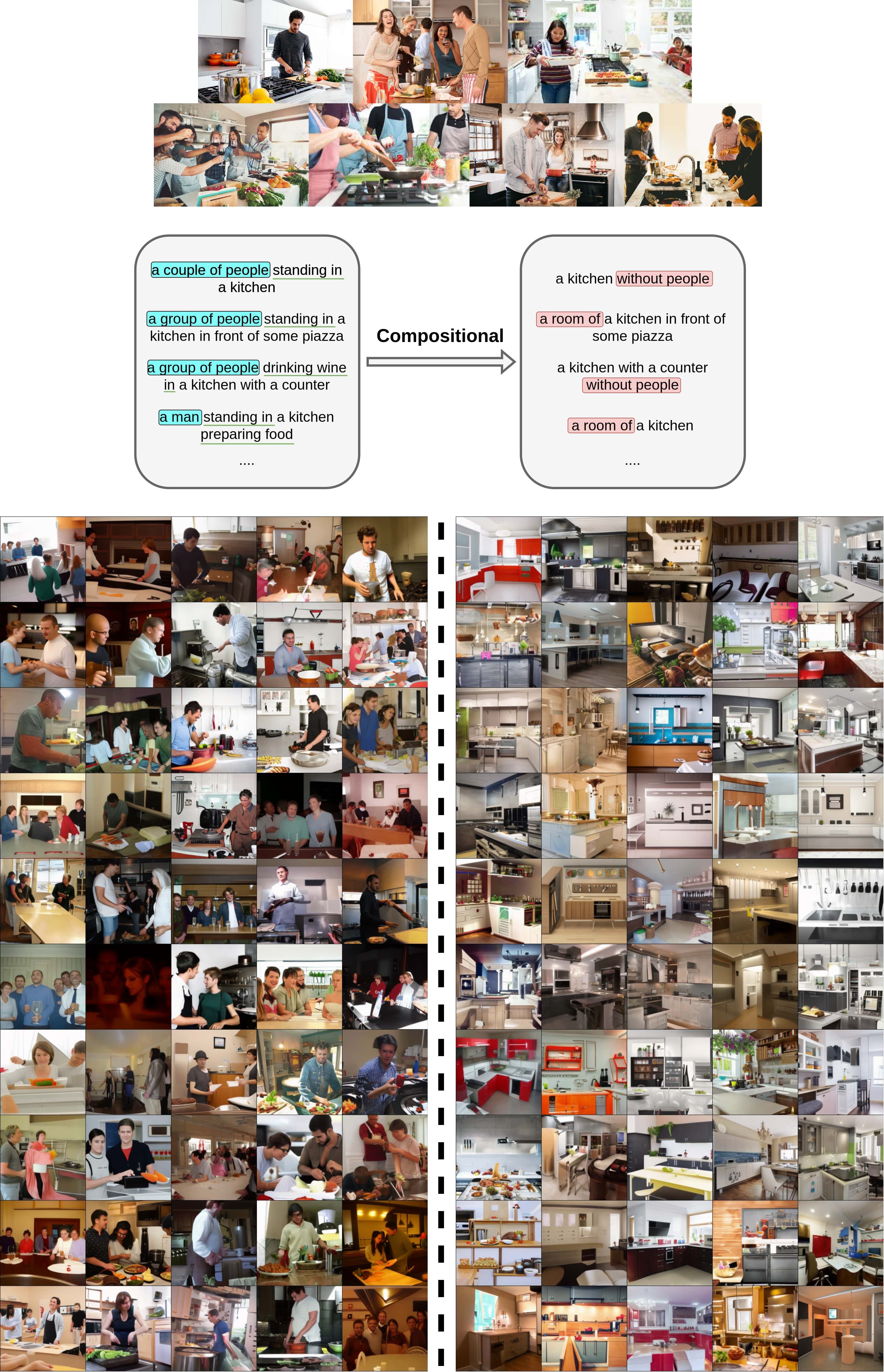}
\end{center}
   \caption{Visualization of context images generated by DALL-E for noisy and out-of-distribution CDIs. In the left column, inputs CDIs have distractor objects (people). The compositional property allow us \textbf{remove} the distractor words from captions and generate high-quality context images (right column). }
\label{fig:human_appendix}
\end{figure*}

\paragraph{Details of GMU kitchen results}

We add per-class details of the performance on GMU kitchen dataset (Table.~\ref{table:gmu-kitchen})

\section{Model Inference Visualization }\label{sec:model_prediction}

\begin{figure*}[h]
\begin{center}
\includegraphics[width=1.0\linewidth]{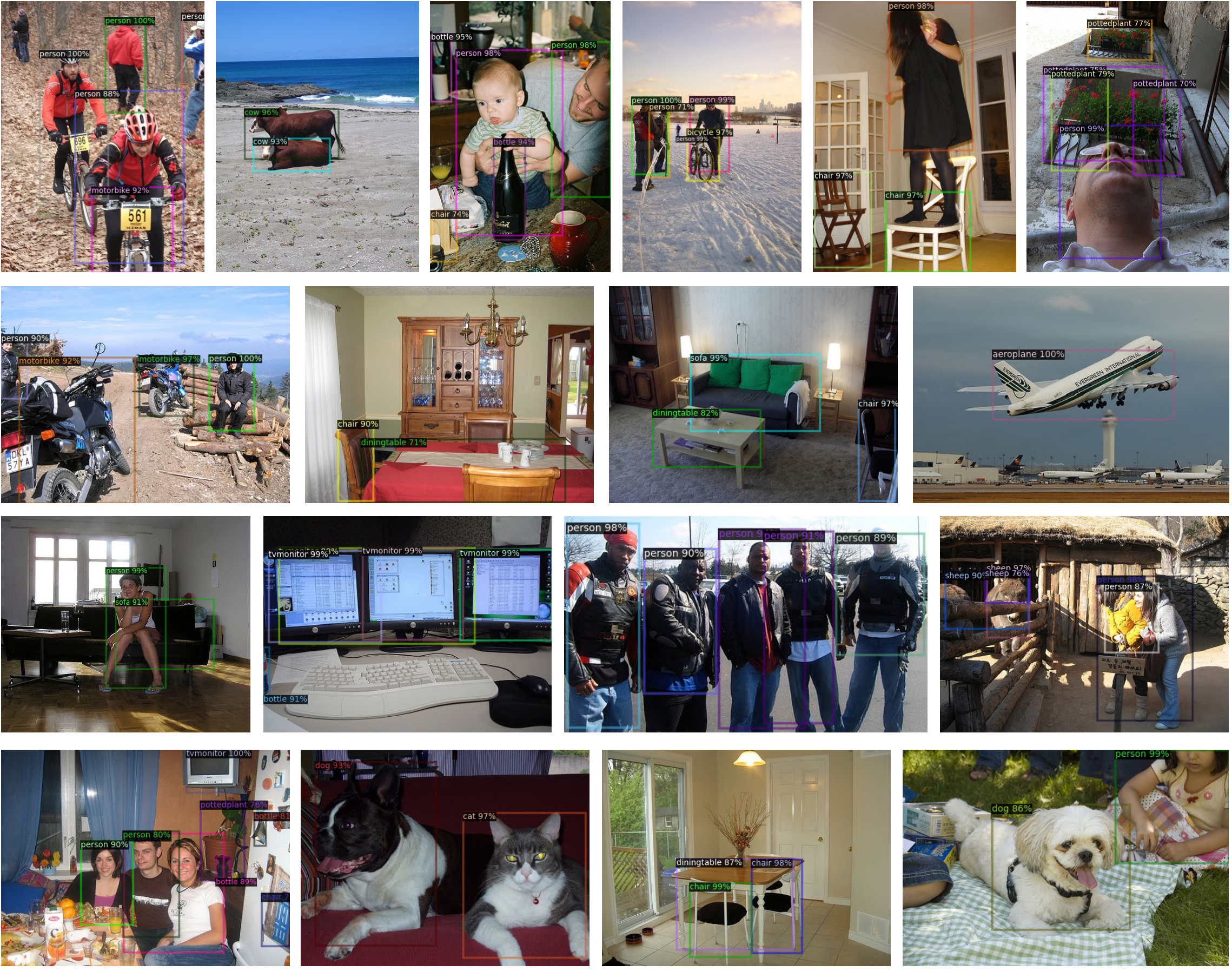}
\end{center}
   \caption{Qualitative detection results on PASCAL VOC dataset. The colorful box is the prediction of the model, together with the predicted class and confidence. We only show prediction with confidence $> 0.9$. The different color of the box indicates the different prediction of the box.}
\label{fig:supp-inference}
\end{figure*}

\begin{figure*}[h]
\begin{center}
\includegraphics[width=1.0\linewidth]{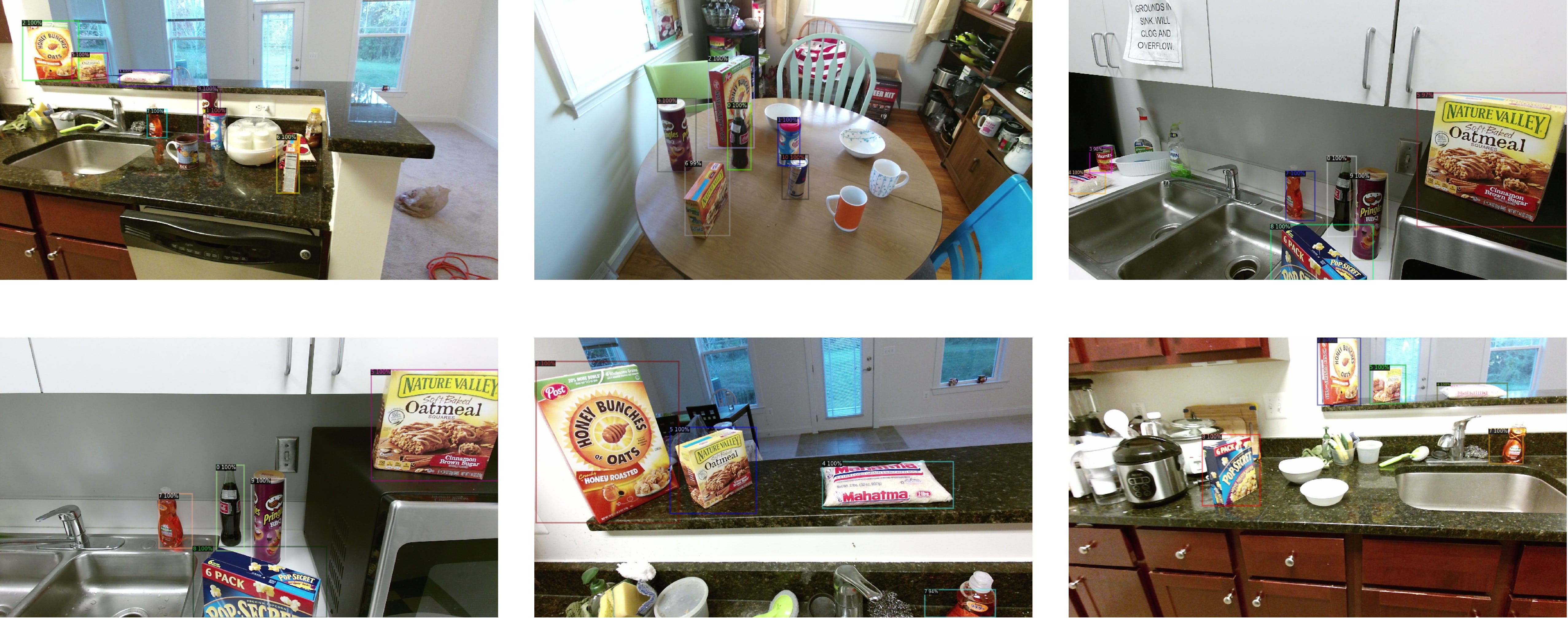}
\end{center}
   \caption{Qualitative detection results on GMU-Kitchen dataset dataset. The colorful box is the prediction of the model, together with the predicted class and confidence. We only show prediction with confidence $> 0.9$. The different color of the box indicates the different prediction of the box.}
\label{fig:detectron_appenfix}
\end{figure*}

Fig.~\ref{fig:supp-inference} demonstrates the predictions of faster RCNN \cite{ren2017faster} models on PASCAL VOC \cite{everingham2010pascal}. The model was trained with the setting of main paper Table 2 G-2 EXP-5.
Fig.~\ref{fig:detectron_appenfix} demonstrates the predictions of faster RCNN on the GMU-Kitchen dataset. The model was trained with \# CDI = 1500 from UW dataset \cite{georgakis2016multiview}. 
Although the model is trained on synthetic data, the model is able to yield good predictions without being accessible to GMU Kitchen's train data.

\end{document}